\documentclass[lettersize,journal]{IEEEtran}

\usepackage{subcaption}
\usepackage{amsmath,amsfonts}
\usepackage{algorithmic}
\usepackage{algorithm}
\usepackage{array}
\usepackage{textcomp}
\usepackage{stfloats}
\usepackage{url}
\usepackage{verbatim}
\usepackage{graphicx}
\usepackage{cite}
\usepackage{multirow}
\usepackage{booktabs}
\usepackage[dvipsnames]{xcolor}
\usepackage{colortbl}
\usepackage{listings}
\usepackage{pifont}
\usepackage{caption}
\usepackage[normalem]{ulem}
\usepackage{adjustbox}

\usepackage[pagebackref=false,colorlinks,urlcolor=magenta]{hyperref}

\definecolor{fr_purple}{RGB}{139,85,155}
\definecolor{fr_green}{RGB}{78,160,88}

\usepackage[capitalize]{cleveref}
\crefname{section}{Sec.}{Secs.}
\Crefname{section}{Section}{Sections}
\Crefname{table}{Table}{Tables}
\crefname{table}{Tab.}{Tabs.}

\begin{document}

\title{FRNet: Frustum-Range Networks for Scalable LiDAR Segmentation}

\author{Xiang Xu, Lingdong Kong,~\IEEEmembership{Student Member,~IEEE}, Hui Shuai, and Qingshan Liu,~\IEEEmembership{Senior Member,~IEEE}
\thanks{
X. Xu is with the College of Computer Science and Technology, Nanjing University of Aeronautics and Astronautics, Nanjing, China.}
\thanks{L. Kong is with the School of Computing, Department of Computer Science, National University of Singapore, Singapore.}
\thanks{H. Shuai and Q. Liu are with the School of Computer Science, Nanjing University of Posts and Telecommunications, Nanjing, China.}
\thanks{The corresponding author is Qingshan Liu: \url{qsliu@njupt.edu.cn}.}
\thanks{This work was supported in part by the Natural Science Foundation of China under Grants U21B2044 and U24B20155, and in part by the Jiangsu Province Science and Technology Project under Grants BA2022026 and BK20243051.}}

\markboth{Journal of \LaTeX\ Class Files,~Vol.~A, No.~B, April~2024}%
{Xu \MakeLowercase{\textit{et al.}}: FRNet: Frustum-Range Networks for Scalable LiDAR Segmentation}

\maketitle
\begin{abstract}
LiDAR segmentation has become a crucial component of advanced autonomous driving systems. Recent range-view LiDAR segmentation approaches show promise for real-time processing. However, they inevitably suffer from corrupted contextual information and rely heavily on post-processing techniques for prediction refinement. In this work, we propose \textbf{FRNet}, a simple yet powerful method aimed at restoring the contextual information of range image pixels using corresponding frustum LiDAR points. First, a frustum feature encoder module is used to extract per-point features within the frustum region, which preserves scene consistency and is critical for point-level predictions. Next, a frustum-point fusion module is introduced to update per-point features hierarchically, enabling each point to extract more surrounding information through the frustum features. Finally, a head fusion module is used to fuse features at different levels for final semantic predictions. Extensive experiments conducted on four popular LiDAR segmentation benchmarks under various task setups demonstrate the superiority of FRNet. Notably, FRNet achieves 73.3\% and 82.5\% mIoU scores on the testing sets of SemanticKITTI and nuScenes. While achieving competitive performance, FRNet operates 5 times faster than state-of-the-art approaches. Such high efficiency opens up new possibilities for more scalable LiDAR segmentation. The code has been made publicly available at \url{https://github.com/Xiangxu-0103/FRNet}.
\end{abstract}


\section{Introduction}
\label{sec:introduction}

\IEEEPARstart{L}{iDAR} segmentation, a crucial and indispensable component in modern autonomous driving, robotics, and other safety-critical applications, has witnessed substantial advancements recently~\cite{guo2020deep,kong2024sc_lpr,li2024place3d,kong2025calib3d}. The principal challenge lies in devising a \textit{scalable} LiDAR segmentation system that strikes a delicate balance between efficiency and accuracy, especially in resource-constrained operational scenarios~\cite{sun2024image,hu2022leveraging}. Here, ``scalable'' refers to the system's ability to maintain high performance while managing increased data loads without compromising real-time processing capabilities.

Existing LiDAR segmentation methods adopt various data representation perspectives, each with inherent trade-offs in terms of accuracy and computational efficiency. Point-based methods~\cite{thomas2019kpconv,hu2020randla,shuai2021baflac,zhang2023pids,puy2023waffleiron,wu2024ptv3} manipulate the original point cloud to preserve spatial granularity but entail computationally expensive neighborhood searches to construct local structures. This demand results in significant computational overhead and limits their capability to deal with large-scale point clouds~\cite{li2022panoptic,hong20244ddsnet,zhang2023pids}. Sparse-voxel-based methods~\cite{tang2020spvnas,zhu2021cylinder3d,choy2019minkunet,zhao2022svaseg,vanian2022improving} transform scattered LiDAR points into regular voxel grids and leverage popular sparse convolutions to extract features. However, this process also demands heavy computations, especially for applications that require high voxel resolutions~\cite{zhu2021cylinder3d,tang2020spvnas,boulch2023also}. Multi-view methods~\cite{liong2020amvnet,xu2021rpvnet,alnaggar2021mpf,liu2023uniseg} extract features from various representations to enhance prediction accuracy~\cite{chen2023clip2scene,chen2023towards,jaritz2022xumda}. While this strategy can improve performance, it aggregates the computation required for all representations. Albeit these representation methods achieved satisfactory performance, their computation overhead hinders them from real-time online applications in real life~\cite{xu2024visual,liu2023uniseg,yu2022dair-v2x,xu2023v2v4real,yu2023v2x-seq}.

\begin{figure}
    \centering
    \includegraphics[width=1.0\linewidth]{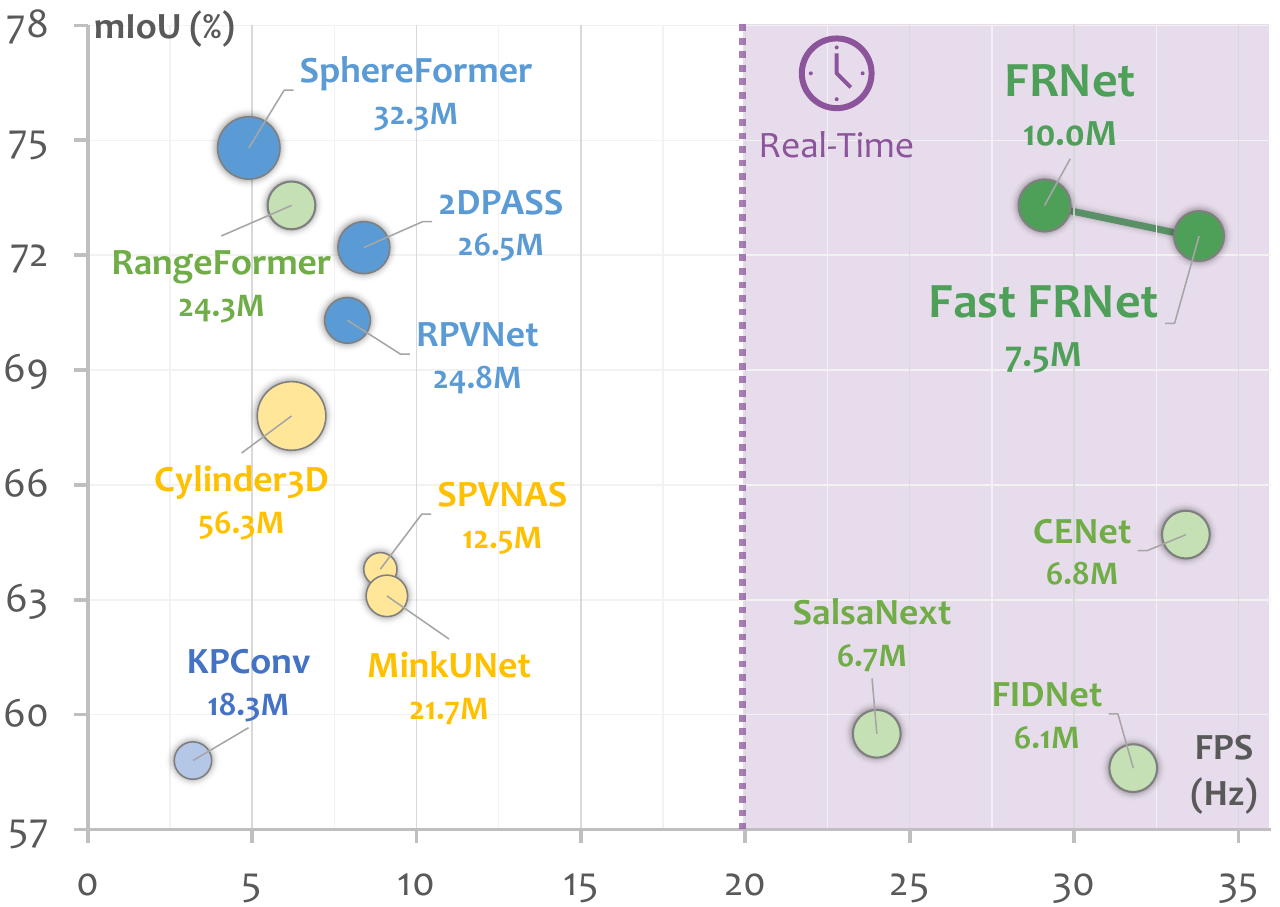}
    \caption{\textbf{A study on the scalability} of state-of-the-art LiDAR segmentation models on the SemanticKITTI~\cite{behley2019semantickitti} leaderboard. The size of the circular representation corresponds to the number of model parameters. FRNet achieves competitive performance with current state-of-the-art models while still maintaining satisfactory efficiency for real-time processing.}
    \label{fig:teaser}
    \vspace{-0.6cm}
\end{figure}

Recently, pseudo-image-based methods~\cite{wu2018squeezeseg,milioto2019rangenet++,ando2023rangevit,kong2023rangeformer,zhang2020polarnet,chen2021polarstream,aksoy2020salsanet,zhou2021panoptic} have appeared as a simple yet efficient intermediate representation for LiDAR processing, such as bird's-eye-view~\cite{zhang2020polarnet,chen2021polarstream,aksoy2020salsanet} and range-view~\cite{wu2018squeezeseg,milioto2019rangenet++,ando2023rangevit,kong2023rangeformer}, which are more computationally tractable. These representations enable the direct application of popular 2D segmentation approaches~\cite{long2015fcn,zhao2017pspnet,xie2021segformer} to the pseudo-images, offering a promising solution for real-time LiDAR segmentation. However, the 3D-to-2D projection inevitably introduces corrupted neighborhood contextual information, which poses limitations on the further development of pseudo-image approaches~\cite{liang2020rangercnn}. The detailed analysis of this problem can be further summarized into two aspects as follows.

Firstly, a fundamental concern with pseudo-image methods is the inadvertent exclusion of certain 3D points during the conversion to 2D projections~\cite{tian2022fully}. This leads to an incomplete representation where only the points that are projected are involved in subsequent convolutional operations. Such a scenario disregards the contextual information of the unprojected points, causing the pseudo-images to lose crucial high-fidelity 3D geometric context~\cite{kong2023conda}. Furthermore, the omitted points, which still necessitate 3D label predictions, lack associated feature information, thereby necessitating post-processing to infer these labels. Unfortunately, current post-processing methods struggle to incorporate local features of these omitted points effectively, capping the potential advancements in pseudo-image approaches.

Secondly, the challenge with pseudo-images is exacerbated by the significant presence of empty pixels due to the inherent sparsity of LiDAR data. For instance, in the nuScenes dataset, nearly 60\% of range image pixels are found to be void~\cite{fong2022panoptic,kong2023rangeformer}. This vacancy not only skews the scene representation but also compromises the quality of semantic segmentation. As the network processes deeper layers, these sparse pseudo-images can become riddled with noise, obstructing the extraction of meaningful patterns. In essence, although pseudo-image methods excel in terms of real-time processing capabilities, their reduced dimensional perspective can overlook crucial 3D details, leading to degraded scalability.

Observing the above issues, in this work, we propose a Frustum-Range Network (FRNet) for scalable LiDAR segmentation, which incorporates points into the range image, achieving a superior balance between efficiency and accuracy. FRNet consists of three main components. Firstly, a Frustum Feature Encoder (FFE) is utilized to group all points with the same frustum region into corresponding range-view pixels in relation to the range image using multiple multi-layer perceptrons (MLPs). This allows for the preservation of all points and the prediction of semantic labels for these points in an end-to-end manner. Subsequently, the point features are pooled to represent the frustum region and formatted into a 2D representation, which is then subjected to traditional convolutions. Secondly, a Frustum-Point (FP) fusion module is employed to efficiently update the hierarchical features of each point during each convolutional stage. This module includes frustum-to-point fusion to update per-point features and point-to-frustum fusion to enhance frustum features. As a result, all points extract larger local features based on the frustum region. Finally, a Fusion Head (FH) module is designed to leverage features from different levels to generate individual features for each point, facilitating end-to-end prediction without the need for post-processing techniques. As shown in \cref{fig:teaser}, the proposed FRNet achieves great improvement among range-view methods while still maintaining high efficiency.

Data augmentation plays an important role in training robust and generalizable models. Inspired by the success of mixing strategies~\cite{kong2023lasermix,xiao2022polarmix,nekrasov2021mix3d}, we introduce a fine-grained mixing method dubbed FrustumMix, which operates based on frustum units. Specifically, we randomly divide point cloud scenes into several frustum regions along inclination and azimuth directions and swap the corresponding regions from different scenes, with the neighborhood of each frustum region coming from the other scene. FrustumMix provides a straightforward and lightweight method to adapt FRNet by focusing on the frustum region. Furthermore, to address the issue of empty pixels in the 2D representation, we propose RangeInterpolation to reconstruct the semantic surface based on surrounding point information. It first transforms the LiDAR point cloud into a range image, as done in previous works~\cite{wu2018squeezeseg,milioto2019rangenet++,ando2023rangevit,kong2023rangeformer}. For empty pixels in the range image, we then aggregate surrounding range information within a pre-defined window using an average pooling operation to generate a new point. RangeInterpolation is able to assist in creating a more compact 2D representation with fewer empty pixels, enabling convolutions to learn more semantically coherent information.

The key contributions of this work are summarized as:
\begin{itemize}
    \item We introduce a novel Frustum-Range representation for scalable LiDAR segmentation. In our framework, the point-level geometric information is integrated into the pseudo-image representation, which not only retains the complete geometric structure of the point cloud, but also takes advantage of the efficiency of the 2D network.
    \item We propose two 3D data augmentation techniques to assist in training robust and generalizable models. FrustumMix generates more complex scenes by mixing two different scans, while RangeInterpolation reconstructs the semantic surfaces based on surrounding range information. These techniques help generate a compact 2D representation and learn more context-aware features.
    \item Extensive experiments across four prevailing LiDAR segmentation benchmarks demonstrate our superiority. FRNet achieves $73.3\%$ and $82.5\%$ mIoU scores, respectively, on SemanticKITTI and nuScenes, while still maintaining promising scalability for real-time LiDAR segmentation. 
\end{itemize}

\section{Related Work}
\label{sec:related_work}

\subsection{Scalable LiDAR Segmentation}

Although point-view~\cite{thomas2019kpconv,hu2020randla,zhang2023pids,puy2023waffleiron,wu2024ptv3}, sparse-voxel-view~\cite{tang2020spvnas,zhu2021cylinder3d,choy2019minkunet,zhao2022svaseg}, and multi-view~\cite{liong2020amvnet,xu2021rpvnet,alnaggar2021mpf,liu2023uniseg,liu2024m3net,liu2023segment} methods have achieved significant success in LiDAR segmentation, they often suffer from high computational costs and are unable to achieve real-time processing. Recently, various methods have been proposed to project LiDAR points into range images along inclination and azimuth directions to balance the efficiency and accuracy of segmentation~\cite{li2022coarse3d}. SqueezeSeg~\cite{wu2018squeezeseg} and SqueezeSegV2~\cite{wu2019squeezesegv2} use the lightweight model SqueezeNet~\cite{iandola2016squeezenet} to retain information. SqueezeSegV3~\cite{xu2020squeezesegv3} introduces a Spatially-Adaptive Convolution module to adjust kernel weights based on the locations of range images. RangeNet++~\cite{milioto2019rangenet++} integrates DarkNet into SqueezeSeg~\cite{wu2018squeezeseg} and proposes an efficient KNN post-processing for segmentation. RangeViT~\cite{ando2023rangevit} and RangeFormer~\cite{kong2023rangeformer} introduce transformer blocks to extract both local and global information from the range image. However, these methods only involve projected points and fail to extract contextual information for discarded points. To address this problem, WaffleIron~\cite{puy2023waffleiron} proposes to mix point and image features with residual connections. Although achieving promising performance, it requires downsampling points based on voxel grid and space neighborhood search for point feature embedding, which hinders it from real-time processing and end-to-end predictions over the entire point cloud. In this work, we propose an efficient integration of point and image features that preserves both the 3D geometric information from raw point features and the efficiency of 2D convolutions, eliminating the need for pre- or post-processing techniques and yielding superior scalability for real-time LiDAR segmentation.

\subsection{LiDAR Data Augmentation}

Recent works have explored various data augmentation techniques for point clouds~\cite{chen2020pointmixup,zhang2022pointcutmix}, which have shown promise in improving performance for indoor scene understanding but exhibit limited generalization to outdoor scenes. Mix3D~\cite{nekrasov2021mix3d} creates new training samples by combining two out-of-context scenes, which incurs higher memory costs. PolarMix~\cite{xiao2022polarmix} proposes to mix two LiDAR scenes based on azimuth angles and crop instances from one scene to another, while LaserMix~\cite{kong2023lasermix} suggests exchanging labeled scans with unlabeled scans along the inclination direction for efficient semi-supervised learning. RangeFormer~\cite{kong2023rangeformer} presents a lightweight augmentation method that operates only on projected points in range images, disregarding occluded points. In this work, we introduce augmentation techniques aimed at training a robust and scalable model for LiDAR segmentation, including FrustumMix and RangeInterpolation.

\subsection{LiDAR Post-Processing}

Prior works have proposed various post-processing techniques to reconstruct the semantic labels of occluded points based on 2D results. SqueezeSeg~\cite{wu2018squeezeseg} employs the traditional conditional random field (CRF) as a recurrent neural network (RNN). RangeNet++~\cite{milioto2019rangenet++} introduces a fast K-Nearest-Neighbor (KNN) search to aggregate final results across the entire point cloud. FIDNet~\cite{zhao2021fidnet} uses a Nearest Label Assignment (NLA) method for label smoothing. RangeFormer~\cite{kong2023rangeformer} divides full point clouds into sub-clouds and infers each subset in a supervised manner. However, these methods heavily rely on 2D results and fail to extract point features over the entire point cloud in a supervised manner. In contrast, KPRNet~\cite{kochanov2020kprnet} and RangeViT~\cite{ando2023rangevit} incorporate a KPConv~\cite{thomas2019kpconv} block to extract point features in an end-to-end manner. Nevertheless, this approach requires substantial computational resources to search neighborhoods in 3D space, which hinders them from real-time processing. In this work, we partition points into frustum regions and extract local features of points in a 2D manner, avoiding time-consuming neighbor searches. Based on the learned point features, FRNet predicts results over the entire point cloud in an end-to-end manner with high efficiency.

\section{FRNet: A Scalable LiDAR Segmentor}
\label{sec:methods}

\begin{figure}
    \centering
    \includegraphics[width=1.0\linewidth]{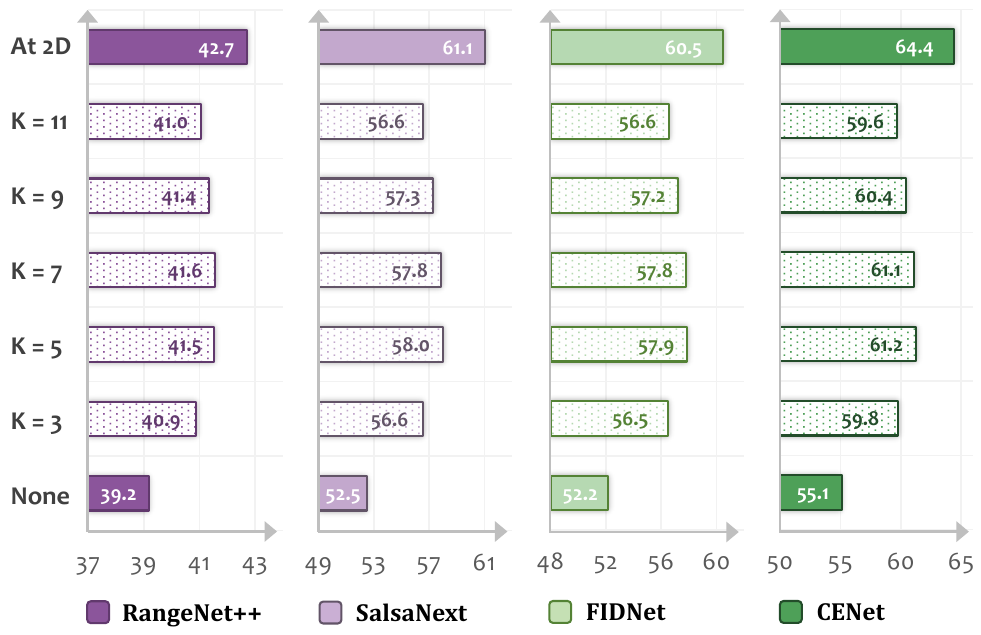}
    \caption{\textbf{Pilot study} on the performance degradation of post-processing in existing range-view methods~\cite{milioto2019rangenet++,aksoy2020salsanet,zhao2021fidnet,cheng2022cenet} on the val set of SemanticKITTI~\cite{behley2019semantickitti}. We choose various $K$ values as hyperparameters in KNN post-processing. Compared to their performance at 2D (\textit{i.e.}, the range image), a severe drop in performance occurs with different $K$ values.}
    \label{fig:KNN}
    \vspace{-0.6cm}
\end{figure}

\begin{figure*}
    \centering
    \includegraphics[width=1.0\linewidth]{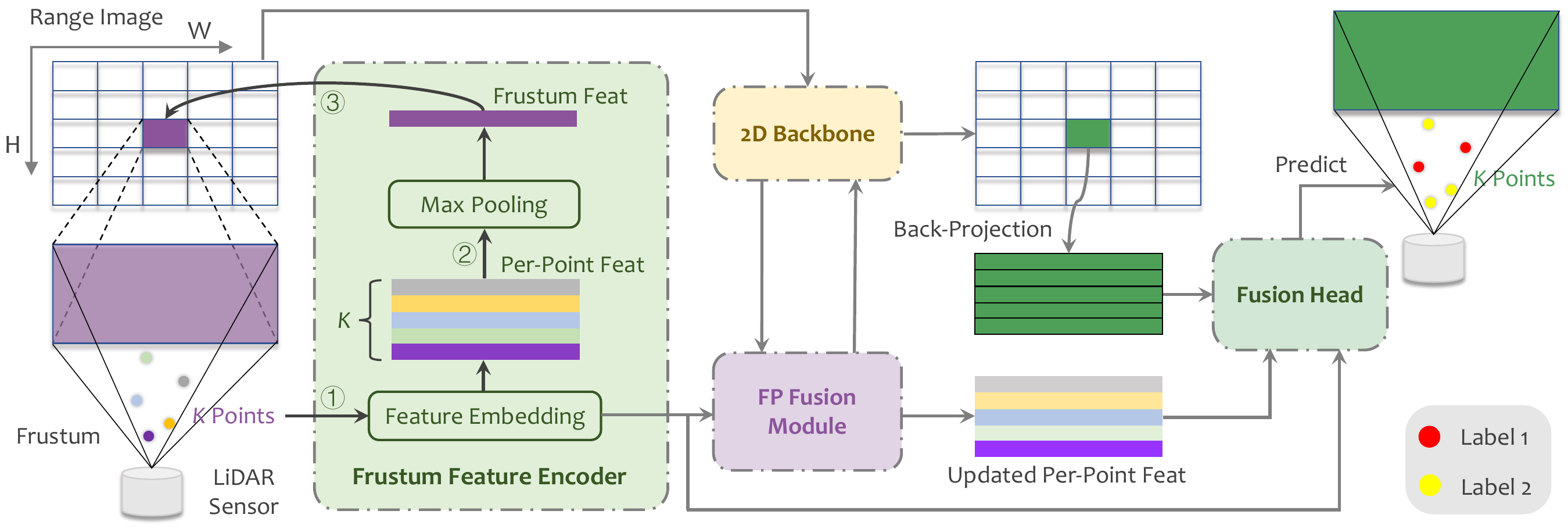}
    \caption{\textbf{Architecture overview}. The proposed FRNet comprises three main components: 1) \textit{Frustum Feature Encoder} is used to embed per-point features within the frustum region. 2) \textit{Frustum-Point (FP) Fusion Module} updates per-point features hierarchically at each stage of the 2D backbone. 3) \textit{Fusion Head} fuses different levels of features to predict final results.}
    \label{fig:framework}
    \vspace{-0.6cm}
\end{figure*}

In this section, we first conduct a pilot study on the most popular range-view LiDAR segmentation methods, wherein we observe a significant impact of KNN post-processing on performance (\cref{sec:pilot_study}). We then elaborate on the technical details of frustum-range architecture (\cref{sec:architecture}) and frustum-range operators (\cref{sec:operator}). The overall pipeline of the proposed FRNet framework is depicted in \cref{fig:framework}.

\subsection{Pilot Study}
\label{sec:pilot_study}

LiDAR segmentation aims to assign semantic labels to each point in the point cloud~\cite{hu2021sensaturban}. A common approach is to operate directly on the LiDAR points without any pre-processing, following the paradigm established by PointNet++~\cite{qi2017pointnet++}. However, point-based operations often require extensive computations and can yield sub-optimal performance due to the sparsity of outdoor scenes. To address this problem, many recent works have proposed projecting LiDAR points onto a regular intermediate representation, known as range images, based on spherical coordinates to enhance efficiency. Specifically, given a point cloud $\mathcal{P}=\{\mathbf{p}_{i}\}_{i=1}^{N}$ with $N$ points, where $\mathbf{p}_{i} \in \mathbb{R}^{3 + L}$ includes the coordinate $(x_{i}, y_{i}, z_{i})$ and additional $L$-dimensional features (intensity, elongation, \textit{etc.}), we map the point to its corresponding 2D position $(u_{i}, v_{i})$ in the range image via the following transformation:
\begin{equation}
    \label{equ:projection}
    \binom{u_{i}}{v_{i}} = \binom{\frac{1}{2}[1 - \arctan(y_{i}, x_{i})\pi^{-1}]W}{[1-(\arcsin(z_{i}d_{i}^{-1}) + \phi_{\text{down}})\phi^{-1}]H}~,
\end{equation}
where $d_{i} = \sqrt{x_{i}^{2} + y_{i}^{2} + z_{i}^{2}}$ is the depth of the point; $\phi = |\phi_{\text{up}}| + |\phi_{\text{down}}|$ represents the vertical field-of-view (FOV) of the sensor, with $\phi_{\text{up}}$ and $\phi_{\text{down}}$ being the inclination angles in the upward and downward directions, respectively; $H$ and $W$ are the height and width of the range image. However, such projection often results in the loss of occluded points and requires post-processing techniques to reconstruct the entire semantic information~\cite{kong2023rangeformer}.

We conducted a pilot study on various range-view methods, including RangeNet++~\cite{milioto2019rangenet++}, SalsaNext~\cite{cortinhal2020salsanext}, FIDNet~\cite{zhao2021fidnet}, and CENet~\cite{cheng2022cenet}. We compared their performances on the original points (\textit{i.e.}, after the 2D-to-3D projection) using the widely-used KNN post-processing with different nearest neighbor parameters, as well as their performance on the 2D range images. As illustrated in \cref{fig:KNN}, the post-processing procedures are often unsupervised, and the choice of hyperparameters can significantly impact the final performance. To address this issue, we propose a scalable Frustum-Range Network (FRNet) capable of directly predicting semantic labels for each point while maintaining high efficiency.

\subsection{Frustum-Range Architecture}
\label{sec:architecture}

\noindent\textbf{Problem Definition.}
Given a point cloud $\mathcal{P}$ acquired by the LiDAR sensor, the objective of FRNet is to employ a feed-forward network $\mathcal{G}$ to predict semantic labels $\mathcal{\widehat{Y}}$ for each individual point as follows:
\begin{equation}\label{equ:formulation}
    \mathcal{\widehat{Y}} = \mathcal{G}(\mathcal{P}, \theta)~,
\end{equation}
where $\theta$ represents the learnable parameters within the network. As depicted in \cref{fig:framework}, FRNet consists of three key components: 1) a frustum feature encoder responsible for per-point feature extraction; 2) a frustum-point fusion module that efficiently fuses hierarchical point features and frustum features; 3) a fusion head module that integrates point features at different levels for accurate semantic prediction.

\noindent\textbf{Operator Representation.}
Let $\text{Flat}(\cdot): \mathbb{R}^{N \times C} \to \mathbb{R}^{H \times W \times C}$ denotes the function that projects point features onto the frustum image plane with resolution $(H, W)$. Within each frustum region, we apply a max-pooling function to obtain the frustum features. Let $\text{MLP}(\cdot): \mathbb{R}^{N \times C} \to \mathbb{R}^{N \times D}$ represents the MLPs that take $C$-dimensional point features as input and output $D$-dimensional point features. Let $\text{Inflat}(\cdot): \mathbb{R}^{H \times W \times C} \to \mathbb{R}^{N \times C}$ be the operation that back-projects frustum features to point features. Notably, points falling into the same frustum region share the same features under this operation.

\noindent\textbf{Frustum Feature Encoder.}
Accurate prediction of semantic labels for each point requires the extraction of individual features. Before converting the point cloud into a 2D frustum representation, the frustum feature encoder plays a pivotal role in extracting per-point features through a series of MLPs, which is crucial for accurate label prediction. The frustum feature encoder follows three steps. \ding{172} We divide the point cloud into $H$ and $W$ frustum regions along the vertical and horizontal directions, respectively. According to \cref{equ:projection}, for each point $\mathbf{p}_{i}$ in the point cloud $\mathcal{P}$, we calculate its frustum position $(u_{i}, v_{i})$. Points sharing the same $(u, v)$ coordinates are grouped into the same frustum region. This results in a set of $M$ frustum regions, defined as $\mathcal{P} = \{\mathcal{P}_{1},\mathcal{P}_{2},...,\mathcal{P}_{M}\}$, where $M = H \times W$ and $\mathcal{P}_{i} \in \mathbb{R}^{N_{i} \times (3 + L)}$ consists of $N_{i}$ points within the $i$-th frustum region. \ding{173} To explicitly embed the frustum structure, we leverage an average pooling function within each frustum region to obtain the cluster points $\mathcal{\widetilde{P}} = \{\mathcal{\widetilde{P}}_{1},\mathcal{\widetilde{P}}_{2},...,\mathcal{\widetilde{P}}_{M}\}$, where $\mathcal{\widetilde{P}}_{i} \in \mathbb{R}^{1 \times (3 + L)}$. Subsequently, we construct a frustum graph between the point cloud $\mathcal{P}$ and the cluster points $\mathcal{\widetilde{P}}$ within each frustum region. The per-point features are embedded as follows:
\begin{equation}
    \label{equ:frustum_encoder}
    \mathcal{F}_{p}^{0} = \text{MLP}([\mathcal{P}; \mathcal{P} - \widetilde{\mathcal{P}}])~,
\end{equation}
where $[~\cdot~;~\cdot~]$ denotes feature concatenation. A max-pooling function is then applied within the frustum region to yield the initialized frustum feature. \ding{174} The frustum feature is transformed into a 2D frustum representation based on its frustum position $(u, v)$: $\mathcal{F}_{f}^{0} = \text{Flat}(\mathcal{F}_{p}^{0})$. Both $\mathcal{F}_{p}^{0}$ and $\mathcal{F}_{f}^{0}$ serve as inputs to our backbone for the efficient hierarchical updating of point and frustum features.

\begin{figure}
    \centering
    \includegraphics[width=1.0\linewidth]{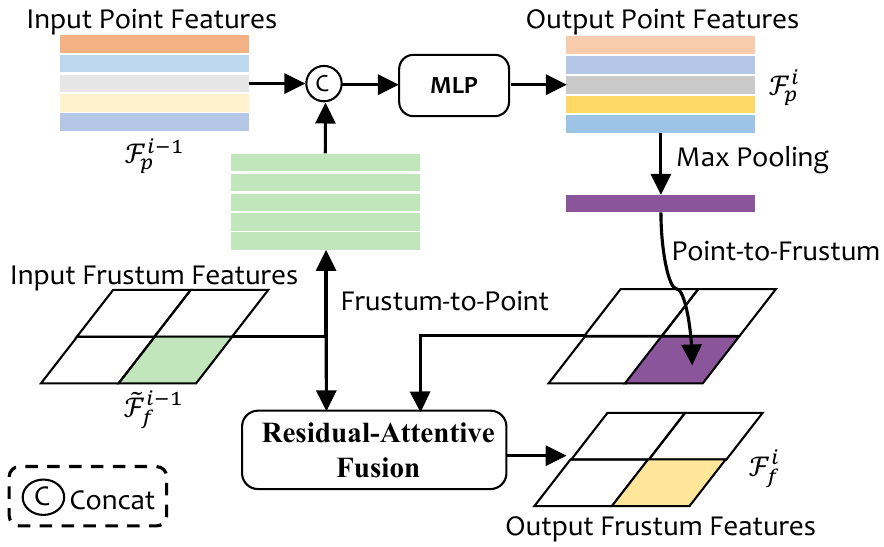}
    \caption{\textbf{Frustum-point fusion module} comprises two steps: 1) A Frustum-to-Point fusion to update per-point features. 2) A Point-to-Frustum fusion to update frustum features.}
    \label{fig:fp_fusion}
    \vspace{-0.6cm}
\end{figure}

\noindent\textbf{Frustum-Point Fusion Module.}
Building on prior works~\cite{zhao2021fidnet,cheng2022cenet,kong2023rangeformer}, we utilize a 2D backbone composed of multiple convolution blocks to efficiently extract hierarchical frustum features. Each stage block is accompanied by a frustum-point fusion module, facilitating the incremental update of point and frustum features. As illustrated in \cref{fig:fp_fusion}, the frustum-point fusion module consists of two essential components: 1) a frustum-to-point fusion that enables the hierarchical update of per-point features, and 2) a point-to-frustum fusion that fuses individual features of each point within its corresponding frustum region. The $i$-th frustum-point fusion module takes $\mathcal{F}_{p}^{i-1}$ and $\mathcal{\widetilde{F}}_{f}^{i-1}$ as inputs, where $\mathcal{\widetilde{F}}_{f}^{i-1}$ is the output of the 2D convolution block that takes $\mathcal{F}_{f}^{i-1}$ as input.

To efficiently extract context information for the points, we employ a frustum-to-point fusion block to integrate local features from the frustum region into the points. First, $\widetilde{\mathcal{F}}_{f}^{i-1}$ is back-projected to the point-level features according to the projection index. Subsequently, the back-projected features are concatenated with $\mathcal{F}_{p}^{i-1}$ and passed through an MLP to update the per-point features, avoiding high computational costs from neighborhood searches:
\begin{equation}
    \label{equ:f2p}
    \mathcal{F}_{p}^{i} = \text{MLP}([\text{Inflat}(\mathcal{\widetilde{F}}_{f}^{i-1}); \mathcal{F}_{p}^{i-1}])~.
\end{equation}

As discussed above, the 2D representation is non-compact due to the sparsity of the point cloud. The convolution block inevitably introduces noisy information for empty frustums, leading to compact feature representations. To preserve the sparsity attributes of the 2D representation, we design a point-to-frustum fusion module to fuse the sparse and dense frustum features. Specifically, the updated point features $\mathcal{F}_{p}^{i}$ are first pooled into the corresponding frustum region to yield a non-compact 2D representation. Subsequently, the non-compact frustum feature is concatenated with the compact frustum feature $\mathcal{\widetilde{F}}_{f}^{i-1}$ and passed through a simple convolutional layer to reduce the number of channels:
\begin{equation}
    \label{equ:p2f}
    \mathcal{F}_{\text{fuse}}^{i} = \text{Conv}([\text{Flat}(\mathcal{F}_{p}^{i}); \mathcal{\widetilde{F}}_{f}^{i-1}])~,
\end{equation}
where $\text{Conv}(\cdot)$ represents a convolution layer with batch normalization and an activation function. Finally, the fused features are utilized to further enhance the frustum features in a residual-attentive manner:
\begin{equation}
    \label{equ:residual-attentive}
    \mathcal{F}_{f}^{i} = \mathcal{\widetilde{F}}_{f}^{i-1} + \sigma(h(\mathcal{F}_{\text{fuse}}^{i}, \theta)) \odot \mathcal{F}_{\text{fuse}}^{i}~,
\end{equation}
where $\sigma$ represents the sigmoid function, $h(\cdot)$ is a linear function with trainable parameters $\theta$, and $\odot$ indicates element-wise multiplication.

\begin{figure*}[t]
    \centering
    \begin{subfigure}[b]{0.31\linewidth}
        \centering
        \includegraphics[width=\linewidth]{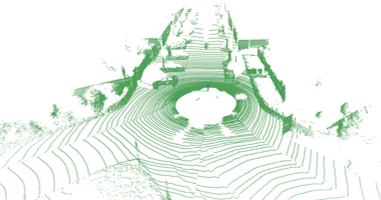}
        \caption{Scene 1}
        \label{fig:scene1}
    \end{subfigure}
    ~~
    \begin{subfigure}[b]{0.31\linewidth}
        \centering
        \includegraphics[width=\linewidth]{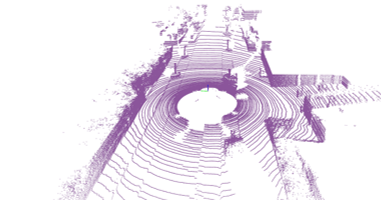}
        \caption{Scene 2}
        \label{fig:scene2}
    \end{subfigure}
    ~~
    \begin{subfigure}[b]{0.31\linewidth}
        \centering
        \includegraphics[width=\linewidth]{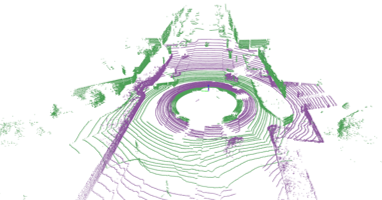}
        \caption{Mixed Scene}
        \label{fig:mixed_scene}
    \end{subfigure}
    \caption{\textbf{FrustumMix illustration.} (a) and (b) show the original two LiDAR scenes. (c) presents the mixed scenes generated by the FrustumMix strategy, where scene 1 is colored \textcolor{fr_green}{green} and scene 2 is colored \textcolor{fr_purple}{purple}.}
    \label{fig:frustum_mix}
    \vspace{-0.2cm}
\end{figure*}

\begin{figure*}[t]
    \centering
    \begin{subfigure}[b]{0.485\linewidth}
        \centering
        \includegraphics[width=\linewidth]{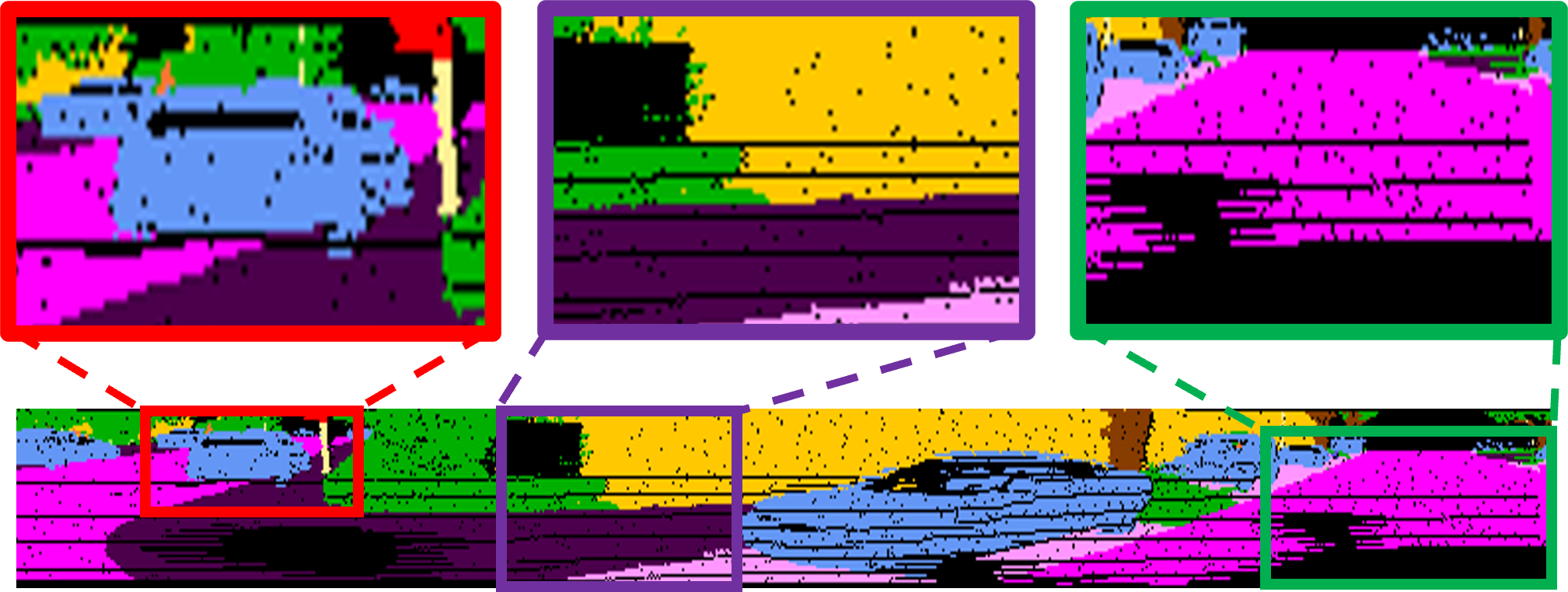}
        \caption{Before RangeInterpolation}
        \label{fig:before_interpolation}
    \end{subfigure}
    ~~
    \begin{subfigure}[b]{0.485\linewidth}
        \centering
        \includegraphics[width=\linewidth]{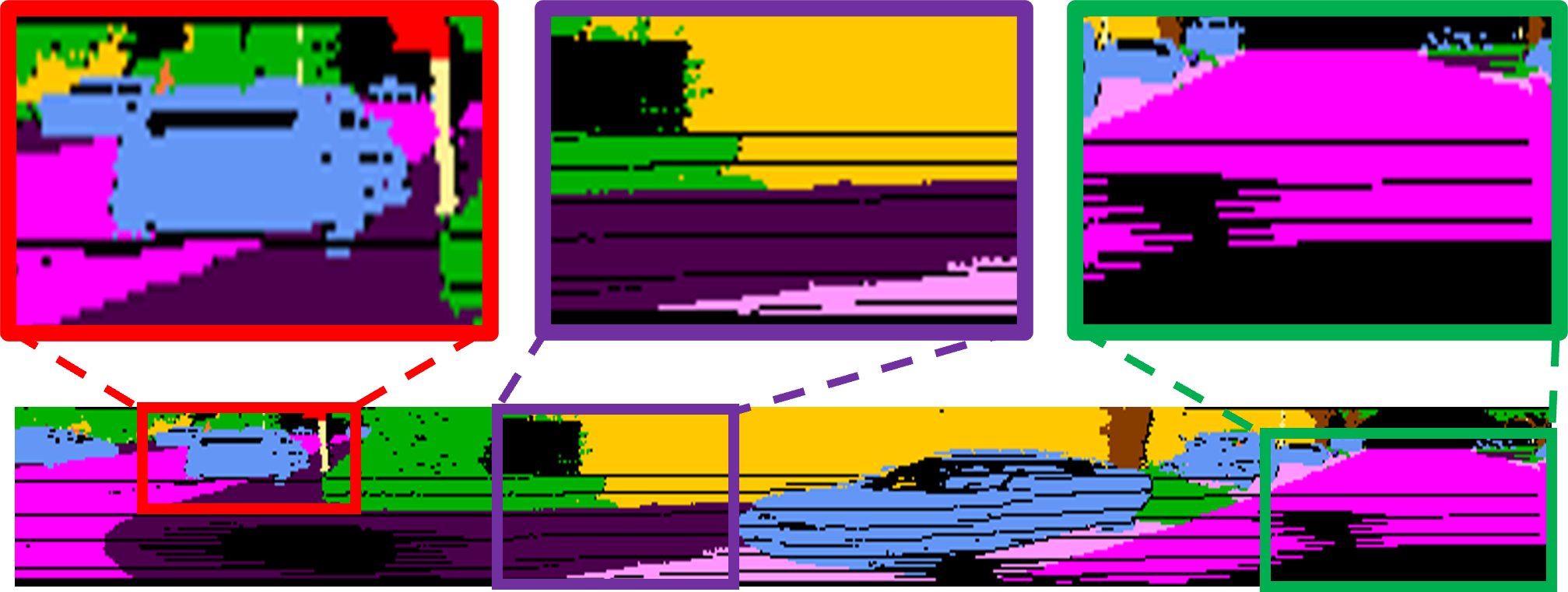}
        \caption{After RangeInterpolation}
        \label{fig:after_interpolation}
    \end{subfigure}
    \caption{\textbf{RangeInterpolation illustration.} (a) shows the raw range image with numerous empty pixels. (b) illustrates the range image after applying RangeInterpolation, resulting in a smoother and more compact representation.}
    \label{fig:range_interpolation}
    \vspace{-0.4cm}
\end{figure*}

\noindent\textbf{Fusion Head Module.}
FRNet aims to assign a category to each point in the point cloud, which requires point-level features for end-to-end label prediction. Although the output point features from the last layer of the frustum-point fusion module can yield promising performance, we observe this is not the optimal solution. Features from different levels have different context-aware information. Lower-level features are geometric while higher-level features are semantically rich. Leveraging these different level features can enhance performance. Drawing inspiration from FIDNet~\cite{zhao2021fidnet} and CENet~\cite{cheng2022cenet}, we first combine all point and frustum features from each stage of the backbone and fuse them to reduce channel dimension:
\begin{equation}
    \label{equ:point_backbone_fusion}
    \mathcal{F}_{p}^{\text{out}} = \text{MLP}([\mathcal{F}_{p}^{1};...;\mathcal{F}_{p}^{V}])~,
\end{equation}
\begin{equation}
    \label{equ:frustum_backbone_fusion}
    \mathcal{F}_{f}^{\text{out}} = \text{Conv}([\mathcal{F}_{f}^{1};...;\mathcal{F}_{f}^{V}])~,
\end{equation}
where $V$ is the number of stages in the backbone. Fusing backbone features with different receptive fields helps obtain more context-aware and semantic information. Additionally, point features $\mathcal{F}_{p}^{\text{out}}$ contains local information from neighborhood frustum regions, while frustum features $\mathcal{F}_{f}^{\text{out}}$ contains global information within the corresponding frustum region. Meanwhile, $\mathcal{F}_{p}^{0}$ from the frustum feature encoder module constructs graph structures within the frustum region and between neighborhood frustum regions. Progressively fusing these features helps extract features from local to global, and from geometric to semantic, resulting in more accurate predictions. To fuse $\mathcal{F}_{f}^{\text{out}}$ and $\mathcal{F}_{p}^{\text{out}}$, we first back-project the frustum features to point-level and use an MLP layer to ensure the same channels as $\mathcal{F}_{p}^{\text{out}}$. Then, an addition operation is applied to fuse the two features. The same operation is utilized to fuse the new features with $\mathcal{F}_{p}^{0}$. The overall operation can be formulated as:
\begin{equation}
    \label{equ:head}
    \mathcal{F}_{\text{logit}} = \text{MLP}(\text{MLP}(\text{Inflat}(\mathcal{F}_{f}^{\text{out}}))+\mathcal{F}_{p}^{\text{out}})+\mathcal{F}_{p}^{0}~.
\end{equation}
Here, $\mathcal{F}_{\text{logit}}$ is used to generate the final semantic scores with a linear head for the point over the entire point cloud.

\noindent\textbf{Optimization.}
In the conventional setup, a point-level loss function is commonly employed to supervise the final semantic scores. However, in this work, we propose optimizing frustum features with additional auxiliary heads, utilizing the frustum-level loss function. This approach requires generating pseudo frustum labels. Typically, a frustum region encompasses multiple points, which may not all belong to the same category. To address this, we adopt a statistical approach inspired by sparse-voxel methods~\cite{zhu2021cylinder3d}. We generate a pseudo label for each frustum by first counting the frequency of each category existing in the given frustum region, and then choosing the category with the highest frequency as the pseudo label for the frustum. Finally, the frustum-level features are supervised using the generated pseudo labels. The overall loss function can be formulated as follows:
\begin{equation}
    \label{equ:loss}
    \mathcal{L} = \mathcal{L}_{p} + \lambda \mathcal{L}_{f}~,
\end{equation}
where $\mathcal{L}_{p}$ and $\mathcal{L}_{f}$ represent the point-level and frustum-level loss functions, respectively, and $\lambda$ is a hyperparameter that controls the weight of frustum supervision. $\mathcal{L}_{p}$ is calculated by cross-entropy loss, while $\mathcal{L}_{f}$ consists of cross-entropy loss, Lov{\'a}sz-Softmax loss~\cite{berman2018lovasz}, and boundary loss~\cite{razani2021boundary}.

\subsection{Frustum-Range Operators}
\label{sec:operator}

Data augmentation enables the model to learn more robust representations. Common augmentations in LiDAR segmentation typically operate from a global perspective, including rotation, flipping, scaling, jittering, \textit{etc}. In this work, we introduce two novel augmentations based on our proposed frustum-range view representation.

\noindent\textbf{FrustumMix.}
Previous mixing strategies~\cite{kong2023lasermix,xiao2022polarmix} perform point-level swapping between two LiDAR scans. While these methods achieve notable improvements in various LiDAR representations, they are not well aligned with the frustum representation, potentially causing two distinct scenes to mix within the same frustum region. This misalignment disrupts both the semantic coherence and geometric structure within the frustum region. To this end, we propose FrustumMix, a fine-grained extension of LaserMix~\cite{kong2023lasermix} that aligns better with the frustum representation. Specifically, given two LiDAR point clouds $\mathcal{P}_{1}$ and $\mathcal{P}_{2}$, we randomly split the point clouds into $M$ non-overlapping frustum regions along the inclination or azimuth direction, denoted as $\mathcal{P}_{1} = \{\mathcal{A}_{1}^{1},...,\mathcal{A}_{1}^{M}\}$ and $\mathcal{P}_{2} = \{\mathcal{A}_{2}^{1},...,\mathcal{A}_{2}^{M}\}$. We then generate a new LiDAR scan $\hat{\mathcal{P}}$ by alternately swapping frustum regions from $\mathcal{P}_{1}$ and $\mathcal{P}_{2}$, as follows: $\hat{\mathcal{P}} = \mathcal{A}_{1}^{1}\cup\mathcal{A}_{2}^{2}\cup\mathcal{A}_{1}^{3}\cup\mathcal{A}_{2}^{4}\cup\cdots$. This approach ensures that mixing occurs at the level of frustum units, preserving semantic coherence and structural details entirely within each frustum region, thereby maintaining alignment with the frustum representation. \cref{fig:frustum_mix} provides an illustration of FrustumMix. By enhancing context-awareness between frustums while preserving intra-frustum structural invariance, FrustumMix achieves improved segmentation performance and better representation learning.

\begin{table*}
    \centering
    \caption{\textbf{Supervised LiDAR segmentation} results on the official SemanticKITTI~\cite{behley2019semantickitti}, nuScenes~\cite{fong2022panoptic}, ScribbleKITTI~\cite{unal2022scribblekitti}, and SemanticPOSS~\cite{pan2020semanticposs} benchmarks. $\dagger$ denotes models that use pretrained weights. \textbf{FPS} denotes frame rate per second ($s$). All \textbf{mIoU} and \textbf{mAcc} scores are given in percentage ($\%$). The \textbf{best} and \underline{second best} scores for models from each representation group are highlighted in \textbf{bold} and \underline{underline}.}
    \resizebox{\linewidth}{!}{
    \begin{tabular}{r|r|r|r|p{18pt}<{\centering}p{18pt}<{\centering}p{18pt}<{\centering}|p{18pt}<{\centering}p{18pt}<{\centering}|p{18pt}<{\centering}p{18pt}<{\centering}|p{18pt}<{\centering}p{18pt}<{\centering}}
        \toprule
        \multirow{2}{*}{\textbf{Method}} & \multirow{2}{*}{\textbf{Venue}} & \multirow{2}{*}{\textbf{Representation}} & \multirow{2}{*}{\textbf{Param}} & \multicolumn{3}{c|}{\textbf{SemanticKITTI}} & \multicolumn{2}{c|}{\textbf{nuScenes}} & \multicolumn{2}{c|}{\textbf{ScribbleKITTI}} & \multicolumn{2}{c}{\textbf{SemanticPOSS}}
        \\
        & & & & \textcolor{darkgray}{\textbf{FPS}~{\small$\uparrow$}} & \textcolor{darkgray}{\textbf{Val}~{\small$\uparrow$}} & \textcolor{darkgray}{\textbf{Test}~{\small$\uparrow$}} & \textcolor{darkgray}{\textbf{Val}~{\small$\uparrow$}} & \textcolor{darkgray}{\textbf{Test}~{\small$\uparrow$}} & \textcolor{darkgray}{\textbf{mIoU}~{\small$\uparrow$}} & \textcolor{darkgray}{\textbf{mAcc}~{\small$\uparrow$}} & \textcolor{darkgray}{\textbf{mIoU}~{\small$\uparrow$}} & \textcolor{darkgray}{\textbf{mAcc}~{\small$\uparrow$}}
        \\\midrule\midrule
        MinkUNet~\cite{choy2019minkunet} & CVPR'19 & Sparse Voxel \textcolor{Tan}{$\bullet$} & $21.7$ M & \underline{$9.1$} & $62.8$ & $63.7$ & $75.8$ & - & $55.0$ & - & $\mathbf{53.1}$ & $\mathbf{68.3}$
        \\
        SPVNAS~\cite{tang2020spvnas} & ECCV'20 & Sparse Voxel \textcolor{Tan}{$\bullet$} & $21.8$ M & $8.9$ & $62.5$ & $66.4$ & $74.4$ & - & \underline{$56.9$} & - & $48.4$ & $61.5$
        \\
        Cylinder3D~\cite{zhu2021cylinder3d} & CVPR'21 & Sparse Voxel \textcolor{Tan}{$\bullet$} & $55.9$ M & $6.2$ & \underline{$65.9$} & \underline{$67.8$} & $\mathbf{76.1}$ & $\mathbf{77.9}$ & $\mathbf{57.0}$ & - & \underline{$52.9$} & \underline{$64.9$}
        \\
        PVKD~\cite{hou2022pvkd} & CVPR'22 & Sparse Voxel \textcolor{Tan}{$\bullet$} & $14.1$ M & $\mathbf{13.2}$ & $\mathbf{66.4}$ & $\mathbf{71.2}$ & \underline{$76.0$} & - & - & - & - & -
        \\\midrule
        KPConv~\cite{thomas2019kpconv} & ICCV'19 & Raw Points \textcolor{Cyan}{$\bullet$} & $18.3$ M & $3.2$ & $61.3$ & $58.8$ & - & - & - & - & - & -
        \\
        RandLA-Net~\cite{hu2020randla} & CVPR'20 & Raw Points \textcolor{Cyan}{$\bullet$} & $1.2$ M & $1.9$ & $57.1$ & $53.9$ & - & - & - & - & - & -
        \\
        PTv2~\cite{wu2022ptv2} & NeurIPS'22 & Raw Points \textcolor{Cyan}{$\bullet$} & $12.8$ M & $4.7$ & \underline{$70.3$} & \underline{$72.6$} & \underline{$80.2$} & \underline{$82.6$} & - & - & - & -
        \\
        WaffleIron~\cite{puy2023waffleiron} & ICCV'23 & Raw Points \textcolor{Cyan}{$\bullet$} & $6.8$ M & \underline{$7.4$} & $68.0$ & $70.8$ & $79.1$ & - & - & - & - & -
        \\
        PTv3~\cite{wu2024ptv3} & CVPR'24 & Raw Points \textcolor{Cyan}{$\bullet$} & $46.2$ M & $\mathbf{20.2}$ & $\mathbf{70.8}$ & $\mathbf{74.2}$ & $\mathbf{80.4}$ & $\mathbf{82.7}$ & - & - & - & -
        \\\midrule
        RPVNet~\cite{xu2021rpvnet} & ICCV'21 & Multi-View \textcolor{fr_green}{$\bullet$} & $24.8$ M & \underline{$7.9$} & $65.5$ & $70.3$ & $77.6$ & - & - & - & - & -
        \\
        2DPASS~\cite{yan20222dpass} & ECCV'22 & Multi-View \textcolor{fr_green}{$\bullet$} & $26.5$ M & \textbf{$8.4$} & $69.3$ & $72.2$ & $\mathbf{79.4}$ & $80.8$ & - & - & - & -
        \\
        SphereFormer~\cite{lai2023sphereformer} & CVPR'23 & Multi-View \textcolor{fr_green}{$\bullet$} & $32.3$ M & $4.9$ & $67.8$ & $74.8$ & \underline{$78.4$} & $81.9$ & - & - & - & -
        \\
        UniSeg~\cite{liu2023uniseg} & ICCV'23 & Multi-View \textcolor{fr_green}{$\bullet$} & $147.6$ M & $6.9$ & \underline{$71.3$} & \underline{$75.2$} & - & \underline{$83.5$} & - & - & - & -
        \\
        TASeg~\cite{wu2024taseg} & CVPR'24 & Multi-View \textcolor{fr_green}{$\bullet$} & $46.7$ M & $6.1$ & $\mathbf{72.7}$ & $\mathbf{76.5}$ & - & $\mathbf{84.6}$ & - & - & - & -
        \\\midrule
        RangeNet++~\cite{milioto2019rangenet++} & IROS'19 & Range View \textcolor{fr_purple}{$\bullet$} & $50.4$ M & $12.0$ & $50.3$ & $52.2$ & $65.5$ & - & $44.6$ & $57.8$ & $30.9$ & -
        \\
        FIDNet~\cite{zhao2021fidnet} & IROS'20 & Range View \textcolor{fr_purple}{$\bullet$} & $6.1$ M & $31.8$ & $58.9$ & $59.5$ & $71.4$ & - & $54.1$ & $65.4$ & $46.4$ & -
        \\
        CENet~\cite{cheng2022cenet} & ICME'22 & Range View \textcolor{fr_purple}{$\bullet$} & $6.8$ M & \underline{$33.4$} & $62.6$ & $64.7$ & $73.3$ & - & $55.7$ & $66.8$ & $50.3$ & -
        \\
        RangeViT$\dagger$~\cite{ando2023rangevit} & CVPR'23 & Range View \textcolor{fr_purple}{$\bullet$} & $23.7$ M & $10.0$ & $60.7$ & $64.0$ & $75.2$ & - & $53.6$ & $66.5$ & - & -
        \\
        RangeFormer$\dagger$~\cite{kong2023rangeformer} & ICCV'23 & Range View \textcolor{fr_purple}{$\bullet$} & $24.3$ M & $6.2$ & \underline{$67.6$} & $\mathbf{73.3}$ & $78.1$ & $80.1$ & \underline{$63.0$} & - & - & -
        \\
        \textbf{Fast-FRNet} & \textbf{Ours} & Frustum-Range \textcolor{fr_purple}{$\bullet$} & \cellcolor{fr_purple!8}$7.5$ M & \cellcolor{fr_purple!8}$\mathbf{33.8}$ & \cellcolor{fr_purple!8}$67.1$ & \cellcolor{fr_purple!8}\underline{$72.5$} & \cellcolor{fr_purple!8}\underline{$78.8$} & \cellcolor{fr_purple!8}\underline{$82.1$} & \cellcolor{fr_purple!8}$62.4$ & \cellcolor{fr_purple!8}\underline{$71.2$} & \cellcolor{fr_purple!8}\underline{$52.5$} & \cellcolor{fr_purple!8}\underline{$67.1$}
        \\
        \textbf{FRNet} & \textbf{Ours} & Frustum-Range \textcolor{fr_purple}{$\bullet$} & \cellcolor{fr_purple!8}$10.0$ M & \cellcolor{fr_purple!8}$29.1$ & \cellcolor{fr_purple!8}$\mathbf{68.7}$ & \cellcolor{fr_purple!8}$\mathbf{73.3}$ & \cellcolor{fr_purple!8}$\mathbf{79.0}$ & \cellcolor{fr_purple!8}$\mathbf{82.5}$ & \cellcolor{fr_purple!8}$\mathbf{63.1}$ & \cellcolor{fr_purple!8}$\mathbf{72.3}$ & \cellcolor{fr_purple!8}$\mathbf{53.5}$ & \cellcolor{fr_purple!8}$\mathbf{68.1}$
        \\\bottomrule
    \end{tabular}}
    \label{tab:benchmark_sup}
    \vspace{-0.2cm}
\end{table*}

\noindent\textbf{RangeInterpolation.}
Previous range-view methods~\cite{cheng2022cenet,zhao2021fidnet,kong2023rangeformer} often encounter a large number of empty pixels due to the sparsity of LiDAR points, resulting in noisy information. To mitigate this problem, we introduce RangeInterpolation, a technique that utilizes the range image to reconstruct surfaces in the LiDAR scan. Specifically, the point cloud $\mathcal{P}$ is first projected onto the range image $\mathcal{R}\in\mathbb{R}^{H \times W \times (3+L)}$. For empty pixel positions $(u, v)$ in the range image $\mathcal{R}$, we establish a window of size $m \times n$ centered at $(u, v)$, where $m$ and $n$ are odd. We aim to interpolate the point falling into $(u, v)$ using the surrounding range information within the pre-defined window. Considering that not all pixels in the window contain valid values, we aggregate information from non-empty pixels within the window for point interpolation using an average operation. During the training procedure, the semantic label of the interpolated point is determined via threshold-guided voting within the window. Specifically, we identify the most frequent category within the window and its frequency. If the frequency is below a threshold, we classify the interpolated point as located at the boundary and assign the label as ignored, which will not contribute to loss calculation. Otherwise, the label is determined by the category with the highest frequency. As shown in \cref{fig:range_interpolation}, RangeInterpolation produces coherent semantic information in the range image, resulting in a more compact representation.

\section{Experiments}
\label{sec:experiments}

In this section, we demonstrate the scalability and robustness of the proposed method. We first introduce the datasets, metrics, and detailed implementations we used. Subsequently, we show that FRNet achieves an optimal trade-off between efficiency and accuracy when compared to state-of-the-art methods across various task setups. Finally, we conduct a series of ablation studies to analyze each component in FRNet.

\begin{table*}[t]
    \centering
    \caption{\textbf{Semi-supervised LiDAR segmentation} results on the SemanticKITTI~\cite{behley2019semantickitti}, nuScenes~\cite{fong2022panoptic}, and ScribbleKITTI~\cite{unal2022scribblekitti} benchmarks, under an annotation quota of $1\%$, $10\%$, $20\%$, and $50\%$, respectively. All \textbf{mIoU} scores are given in percentage ($\%$). The \textbf{best} and \underline{second best} scores for each data split are highlighted in \textbf{bold} and \underline{underline}.}
    \resizebox{\linewidth}{!}{
    \begin{tabular}{r|r|c|p{15.6pt}<{\centering}p{15.6pt}<{\centering}p{15.6pt}<{\centering}p{15.6pt}<{\centering}|p{15.6pt}<{\centering}p{15.6pt}<{\centering}p{15.6pt}<{\centering}p{15.6pt}<{\centering}|p{15.6pt}<{\centering}p{15.6pt}<{\centering}p{15.6pt}<{\centering}p{15.6pt}<{\centering}}
        \toprule
        \multirow{2}{*}{\textbf{Method}} & \multirow{2}{*}{\textbf{Venue}} & \multirow{2}{*}{\textbf{Backbone}} & \multicolumn{4}{c|}{\textbf{SemanticKITTI}} & \multicolumn{4}{c|}{\textbf{nuScenes}} & \multicolumn{4}{c}{\textbf{ScribbleKITTI}}
        \\
        & & & \textcolor{darkgray}{$\mathbf{1\%}$} & \textcolor{darkgray}{$\mathbf{10\%}$} & \textcolor{darkgray}{$\mathbf{20\%}$} & \textcolor{darkgray}{$\mathbf{50\%}$} & \textcolor{darkgray}{$\mathbf{1\%}$} & \textcolor{darkgray}{$\mathbf{10\%}$} & \textcolor{darkgray}{$\mathbf{20\%}$} & \textcolor{darkgray}{$\mathbf{50\%}$} & \textcolor{darkgray}{$\mathbf{1\%}$} & \textcolor{darkgray}{$\mathbf{10\%}$} & \textcolor{darkgray}{$\mathbf{20\%}$} & \textcolor{darkgray}{$\mathbf{50\%}$}
        \\\midrule\midrule
        \textit{Sup.-only} & - & \multirow{2}{*}{FIDNet~\cite{zhao2021fidnet}} & \cellcolor{fr_green!8}$36.2$ & \cellcolor{fr_green!8}$52.2$ & \cellcolor{fr_green!8}$55.9$ & \cellcolor{fr_green!8}$57.2$ & \cellcolor{fr_green!8}$38.3$ & \cellcolor{fr_green!8}$57.5$ & \cellcolor{fr_green!8}$62.7$ & \cellcolor{fr_green!8}$67.6$ & \cellcolor{fr_green!8}$33.1$ & \cellcolor{fr_green!8}$47.7$ & \cellcolor{fr_green!8}$49.9$ & \cellcolor{fr_green!8}$52.5$
        \\
        LaserMix~\cite{kong2023lasermix} & CVPR'23 & & $43.4$ & $58.8$ & $59.4$ & $61.4$ & $49.5$ & $68.2$ & $70.6$ & $73.0$ & $38.3$ & $54.4$ & $55.6$ & $58.7$
        \\\midrule
        \textit{Sup.-only} & - & \multirow{6}{*}{Cylinder3D~\cite{zhu2021cylinder3d}} & \cellcolor{fr_green!8}$45.4$ & \cellcolor{fr_green!8}$56.1$ & \cellcolor{fr_green!8}$57.8$ & \cellcolor{fr_green!8}$58.7$ & \cellcolor{fr_green!8}$50.9$ & \cellcolor{fr_green!8}$65.9$ & \cellcolor{fr_green!8}$66.6$ & \cellcolor{fr_green!8}$71.2$ & \cellcolor{fr_green!8}$39.2$ & \cellcolor{fr_green!8}$48.0$ & \cellcolor{fr_green!8}$52.1$ & \cellcolor{fr_green!8}$53.8$
        \\
        CRB~\cite{unal2022scribblekitti} & CVPR'22 & & - & $58.7$ & $59.1$ & $60.9$ & - & - & - & - & - & $54.2$ & $56.5$ & $58.9$
        \\
        LaserMix~\cite{kong2023lasermix} & CVPR'23 & & $50.6$ & $60.0$ & $61.9$ & $62.3$ & $55.3$ & $69.9$ & $71.8$ & $73.2$ & $44.2$ & $53.7$ & $55.1$ & $56.8$
        \\
        LiM3D~\cite{li2023lim3d} & CVPR'23 & & - & $61.6$ & $62.6$ & $62.8$ & - & - & - & - & - & $\mathbf{60.3}$ & $\mathbf{60.5}$ & $60.9$
        \\
        ImageTo360~\cite{reichardt2023imageto360} & ICCVW'23 & & $54.5$ & $58.6$ & $61.4$ & $64.2$ & - & - & - & - & - & - & - & -
        \\
        \textbf{FrustumMix} & \textbf{Ours} & & \cellcolor{fr_purple!8}\underline{$55.7$} & \cellcolor{fr_purple!8}$62.5$ & \cellcolor{fr_purple!8}$63.0$ & \cellcolor{fr_purple!8}$64.9$ & \cellcolor{fr_purple!8}$60.0$ & \cellcolor{fr_purple!8}$70.0$ & \cellcolor{fr_purple!8}$72.6$ & \cellcolor{fr_purple!8}$74.1$ & \cellcolor{fr_purple!8}$45.6$ & \cellcolor{fr_purple!8}$55.7$ & \cellcolor{fr_purple!8}$58.2$ & \cellcolor{fr_purple!8}$60.8$
        \\\midrule
        \textit{Sup.-only} & - & \multirow{4}{*}{PTv3~\cite{wu2024ptv3}} & \cellcolor{fr_green!8}$42.7$ & \cellcolor{fr_green!8}$60.9$ & \cellcolor{fr_green!8}$62.6$ & \cellcolor{fr_green!8}$64.1$ & \cellcolor{fr_green!8}$48.0$ & \cellcolor{fr_green!8}$67.8$ & \cellcolor{fr_green!8}$71.5$ & \cellcolor{fr_green!8}$75.9$ & \cellcolor{fr_green!8}$40.7$ & \cellcolor{fr_green!8}$54.6$ & \cellcolor{fr_green!8}$56.2$ & \cellcolor{fr_green!8}$58.9$
        \\
        PolarMix~\cite{xiao2022polarmix} & NeurIPS'22 & & $47.2$ & $61.6$ & $63.1$ & $64.8$ & $53.4$ & $69.3$ & $72.0$ & $76.1$ & $42.5$ & $55.8$ & $56.9$ & $59.6$
        \\
        LaserMix~\cite{kong2023lasermix} & CVPR'23 & & $51.7$ & $63.3$ & $64.8$ & $65.3$ & $57.4$ & $71.8$ & $74.2$ & $77.4$ & $45.0$ & $56.9$ & $58.3$ & $60.7$
        \\
        \textbf{FrustumMix} & \textbf{Ours} & & \cellcolor{fr_purple!8}$54.2$ & \cellcolor{fr_purple!8}\underline{$64.7$} & \cellcolor{fr_purple!8}$\mathbf{65.6}$ & \cellcolor{fr_purple!8}$\mathbf{66.1}$ & \cellcolor{fr_purple!8}\underline{$60.3$} & \cellcolor{fr_purple!8}$\mathbf{72.3}$ & \cellcolor{fr_purple!8}$\mathbf{75.6}$ & \cellcolor{fr_purple!8}$\mathbf{77.7}$ & \cellcolor{fr_purple!8}\underline{$46.2$} & \cellcolor{fr_purple!8}\underline{$57.5$} & \cellcolor{fr_purple!8}\underline{$60.0$} & \cellcolor{fr_purple!8}$\mathbf{61.6}$
        \\\midrule
        \textit{Sup.-only} & - & \multirow{4}{*}{\textbf{FRNet}} & \cellcolor{fr_green!8}$44.9$ & \cellcolor{fr_green!8}$60.4$ & \cellcolor{fr_green!8}$61.8$ & \cellcolor{fr_green!8}$63.1$ & \cellcolor{fr_green!8}$51.9$ & \cellcolor{fr_green!8}$68.1$ & \cellcolor{fr_green!8}$70.9$ & \cellcolor{fr_green!8}$74.6$ & \cellcolor{fr_green!8}$42.4$ & \cellcolor{fr_green!8}$53.5$ & \cellcolor{fr_green!8}$55.1$ & \cellcolor{fr_green!8}$57.0$
        \\
        PolarMix~\cite{xiao2022polarmix} & NeurIPS'22 & & $50.1$ & $60.9$ & $62.0$ & $63.8$ & $55.6$ & $69.6$ & $71.0$ & $73.8$ & $43.2$ & $55.0$ & $56.1$ & $57.3$
        \\
        LaserMix~\cite{kong2023lasermix} & CVPR'23 & & $52.9$ & $62.9$ & $63.2$ & $65.0$ & $58.7$ & $71.5$ & $72.3$ & $75.0$ & $45.8$ & $56.8$ & $57.7$ & $59.0$
        \\
        \textbf{FrustumMix} & \textbf{Ours} & & \cellcolor{fr_purple!8}$\mathbf{55.8}$ & \cellcolor{fr_purple!8}$\mathbf{64.8}$ & \cellcolor{fr_purple!8}\underline{$65.2$} & \cellcolor{fr_purple!8}\underline{$65.4$} & \cellcolor{fr_purple!8}$\mathbf{61.2}$ & \cellcolor{fr_purple!8}\underline{$72.2$} & \cellcolor{fr_purple!8}\underline{$74.6$} & \cellcolor{fr_purple!8}\underline{$75.4$} & \cellcolor{fr_purple!8}$\mathbf{46.6}$ & \cellcolor{fr_purple!8}$57.0$ & \cellcolor{fr_purple!8}$59.5$ & \cellcolor{fr_purple!8}\underline{$61.2$}
        \\\bottomrule
    \end{tabular}}
    \label{tab:benchmark_semi}
    \vspace{-0.4cm}
\end{table*}

\begin{figure*}[t]
    \centering
    \includegraphics[width=1.0\linewidth]{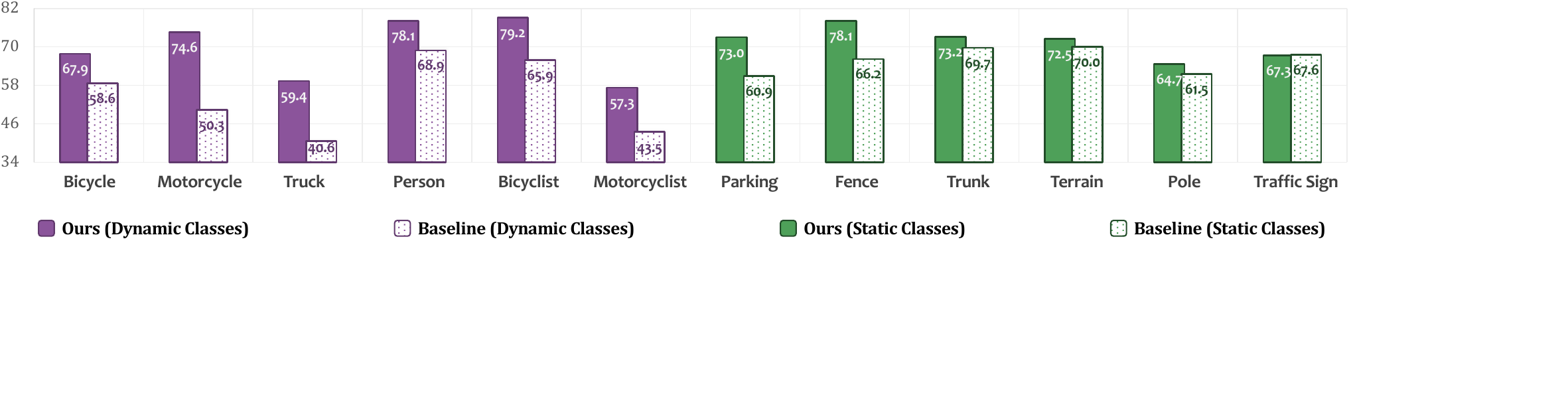}
    \caption{\textbf{Class-wise LiDAR segmentation results} of FRNet and the baseline model on the \textit{test} set of SemanticKITTI~\cite{behley2019semantickitti}.}
    \label{fig:classwise}
    \vspace{-0.4cm}
\end{figure*}

\subsection{Experimental Settings}

\noindent\textbf{Datasets.}
We conduct comprehensive experiments on four popular LiDAR segmentation datasets. \textbf{\textit{SemanticKITTI}}~\cite{behley2019semantickitti} is a large-scale outdoor dataset that comprises $22$ sequences collected from various scenes in Karlsruhe, Germany, using a Velodyne HDL-64E LiDAR sensor. Officially, sequences $00$ to $07$ and $09$ to $10$ ($19,130$ scans) are used for training, sequence $08$ ($4,071$ scans) for validation, and sequences $11$ to $21$ ($20,351$ scans) for online testing. The original dataset is annotated with $28$ classes, though only $19$ merged classes are used for single-scan evaluation. The vertical FOV is $[-25^{\circ}, 3^{\circ}]$. \textbf{\textit{nuScenes}}~\cite{fong2022panoptic} is a multimodal dataset that is widely used in autonomous driving scenarios. It contains LiDAR point clouds captured around streets in Boston and Singapore, provided by 32-beam LiDAR sensors. The dataset includes $1,000$ driving scenes with sparser points. It is annotated with $32$ classes and only $16$ semantic categories are used for evaluation. The vertical FOV is $[-30^{\circ}, 10^{\circ}]$. \textbf{\textit{SemanticPOSS}}~\cite{pan2020semanticposs} is a more challenging benchmark collected at Peking University and features much smaller and sparser point clouds. It consists of $2,988$ LiDAR scenes with numerous dynamic instances. Officially, it is divided into $6$ sequences, with sequence $2$ used for evaluation with $13$ merged categories. The vertical FOV is $[-16^{\circ}, 7^{\circ}]$. \textit{\textbf{ScribbleKITTI}}~\cite{unal2022scribblekitti} shares the same scenes with SemanticKITTI~\cite{behley2019semantickitti} but only provides weak annotations with line scribbles. It contains $189$ million labeled points, with approximately $8.06\%$ labeled points available during training. Additionally, we conduct a comprehensive out-of-distribution robustness evaluation experiment on \textbf{\textit{SemanticKITTI-C}} and \textbf{\textit{nuScenes-C}} from the Robo3D~\cite{kong2023robo3d} benchmark. Each dataset contains eight corruption types, including ``fog'', ``wet ground'', ``snow'', ``motion blur'', ``beam missing'', ``crosstalk'', ``incomplete echo'', and ``cross sensor''.

\noindent\textbf{Evaluation Metrics.}
In line with standard protocols, we utilize Intersection-over-Union (IoU) and Accuracy (Acc) for each class $i$ and compute the mean Intersection-over-Union (mIoU) and mean Accuracy (mAcc) to assess performance. IoU and Acc can be calculated as follows:
\begin{equation}
\text{IoU}_{i} = \frac{\text{TP}_{i}}{\text{TP}_{i} + \text{FP}_{i} + \text{FN}_{i}}~,
\text{Acc}_{i} = \frac{\text{TP}_{i}}{\text{TP}_{i} + \text{FP}_{i}}~,
\end{equation}
where $\text{TP}_{i}, \text{FP}_{i}, \text{FN}_{i}$ are true-positives, false-positives, and false-negatives for class $i$, respectively. For robustness evaluation, we follow the practice of Robo3D~\cite{kong2023robo3d} to use Corruption Error (CE) and Resilience Rate (RR) to evaluate the robustness of FRNet. The CE and RR are calculated as follows:
\begin{equation}
\text{CE}_{i} = \frac{\sum_{l=1}^{3}(1-\text{IoU}_{i,l})}{\sum_{l=1}^{3}(1-\text{IoU}_{i,l}^{\text{base}})}~,
\text{RR}_{i} = \frac{\sum_{l=1}^{3}\text{IoU}_{i,l}}{3 \times \text{IoU}_{\text{clean}}}~,
\end{equation}
where $\text{IoU}_{i,l}$ represents the task-specific IoU. $\text{IoU}_{i,l}^{\text{base}}$ and $\text{IoU}_{\text{clean}}$ denote scores of the baseline model and scores on the ``clean" evaluation set, respectively.

\noindent\textbf{Implementation Details.}
We implement FRNet using the popular MMDetection3D~\cite{mmdet3d} codebase and employ CENet~\cite{cheng2022cenet} as the 2D backbone to extract frustum features. For SemanticKITTI~\cite{behley2019semantickitti} and ScribbelKITTI~\cite{unal2022scribblekitti}, the point cloud is divided into $64 \times 512$ frustum regions, while for nuScenes~\cite{fong2022panoptic} and SemanticPOSS~\cite{pan2020semanticposs}, it is divided into $32 \times 480$. For optimization, we choose AdamW~\cite{loshchilov2019adamw} as our default optimizer with an initial learning rate of $0.01$. We utilized the OneCycle scheduler~\cite{smith2019onecycle} to dynamically adjust the learning rate during training. All models are trained using four GPUs for $50$ epochs on SemanticKITTI~\cite{behley2019semantickitti} and ScribbelKITTI~\cite{unal2022scribblekitti}, and for $80$ epochs on nuScenes~\cite{fong2022panoptic} and SemanticPOSS~\cite{pan2020semanticposs}, respectively. The batch size is set to $4$ for each GPU. Additionally, we implement a faster variant, dubbed Fast-FRNet, which features a larger frustum area and a scaled-down 2D backbone. Specifically, the point cloud is divided into $32 \times 360$ frustum regions for all datasets, and the 2D backbone is a variant of ResNet-18~\cite{he2016resnet}, following the design of CENet~\cite{cheng2022cenet}.

\subsection{Comparative Study}

\noindent\textbf{Comparisons to State of the Arts.}
We begin by comparing FRNet with state-of-the-art LiDAR segmentation methods across different representations, including sparse-voxel, point-view, multi-view, and range-view representations, on both fully-supervised and weakly-supervised benchmarks. \cref{tab:benchmark_sup} provides a summary of all the results. In this benchmark, both FrustumMix and RangeInterpolation are employed as data augmentation techniques to enhance model performance. To ensure a fair comparison, we evaluate the Frame Per Second (FPS) on a single GeForce RTX 2080Ti GPU for all models. For SemanticKITTI~\cite{behley2019semantickitti} and nuScenes~\cite{fong2022panoptic}, we report mIoU scores on both the validation and test sets. For ScribbleKITTI~\cite{unal2022scribblekitti} and SemanticPOSS~\cite{pan2020semanticposs}, which share the same data across the val and test sets, we report both mIoU and mAcc scores on the val set. Compared with recent multi-view methods~\cite{liu2023uniseg}, although FRNet does not exhibit a significant advantage in performance, it shows notable superiority in terms of parameters and FPS. In comparison with range-view, sparse-voxel, and raw-points methods, FRNet achieves appealing performance while still maintaining high efficiency. Additionally, Fast-FRNet demonstrates the fastest inference speed with only a minor drop in performance.

\begin{table*}[t]
    \centering
    \caption{\textbf{Robustness probing} on SemanticKITTI-C and nuScenes-C benchmarks~\cite{kong2023robo3d}. All \textbf{mCE}, \textbf{mRR} and \textbf{mIoU} scores are given in percentage ($\%$). The \textbf{best} and \underline{second best} scores for each model are highlighted in \textbf{bold} and \underline{underline}.}
    \resizebox{\linewidth}{!}{
    \begin{tabular}{c|r|r|p{25pt}<{\centering}|p{25pt}<{\centering}|p{25pt}<{\centering}|p{22.2pt}<{\centering}p{22.2pt}<{\centering}p{22.2pt}<{\centering}p{22.2pt}<{\centering}p{22.2pt}<{\centering}p{22.2pt}<{\centering}p{22.2pt}<{\centering}p{22.2pt}<{\centering}}
        \toprule
        \textbf{\#} & \textbf{Method} & \textbf{Venue} & \textbf{mCE}~{\small$\downarrow$} & \textbf{mRR}~{\small$\uparrow$} & \textbf{mIoU}~{\small$\uparrow$} & \textbf{Fog} & \textbf{Weg} & \textbf{Snow} & \textbf{Motion} & \textbf{Beam} & \textbf{Cross} & \textbf{Echo} & \textbf{Sensor}
        \\\midrule\midrule
        \multirow{8}{*}{\rotatebox[origin=c]{90}{\textbf{SemanticKITTI-C}}} & MinkUNet~\cite{choy2019minkunet} & CVPR'19 & \underline{$100.0$} & \underline{$81.9$} & $62.8$ & $\mathbf{55.9}$ & $54.0$ & $53.3$ & $32.9$ & $56.3$ & $\mathbf{58.3}$ & $54.4$ & $46.1$
        \\
        & SPVCNN~\cite{tang2020spvnas} & ECCV'20 & $100.3$ & $\mathbf{82.2}$ & $62.5$ & \underline{$55.3$} & $54.0$ & $51.4$ & $34.5$ & $56.7$ & \underline{$58.1$} & $54.6$ & $46.0$
        \\
        & FIDNet~\cite{zhao2021fidnet} & IROS'20 & $113.8$ & $77.0$ & $58.8$ & $43.7$ & $51.6$ & $49.7$ & $40.4$ & $49.3$ & $49.5$ & $48.2$ & $29.9$
        \\
        & Cylinder3D~\cite{zhu2021cylinder3d} & CVPR'21 & $103.3$ & $80.1$ & $63.4$ & $37.1$ & $57.5$ & $46.9$ & $52.5$ & $57.6$ & $56.0$ & $52.5$ & $46.2$
        \\
        & 2DPASS~\cite{yan20222dpass} & ECCV'22 & $106.1$ & $77.5$ & $64.6$ & $40.5$ & \underline{$60.7$} & $48.5$ & $\mathbf{57.8}$ & $58.8$ & $28.5$ & $55.8$ & \underline{$50.0$}
        \\
        & CENet~\cite{cheng2022cenet} & ICME'22 & $103.4$ & $81.3$ & $62.6$ & $42.7$ & $57.3$ & \underline{$53.6$} & $52.7$ & $55.8$ & $45.4$ & $53.4$ & $45.8$
        \\
        & WaffleIron~\cite{puy2023waffleiron} & ICCV'23 & $109.5$ & $72.2$ & \underline{$66.0$} & $45.5$ & $58.6$ & $49.3$ & $33.0$ & \underline{$59.3$} & $22.5$ & $\mathbf{58.6}$ & $\mathbf{54.6}$
        \\
        & \cellcolor{fr_purple!8}\textbf{FRNet} & \cellcolor{fr_purple!8}\textbf{Ours} & \cellcolor{fr_purple!8}$\mathbf{96.8}$ & \cellcolor{fr_purple!8}$80.0$ & \cellcolor{fr_purple!8}$\mathbf{67.6}$ & \cellcolor{fr_purple!8}$47.6$ & \cellcolor{fr_purple!8}$\mathbf{62.2}$ & \cellcolor{fr_purple!8}$\mathbf{57.1}$ & \cellcolor{fr_purple!8}\underline{$56.8$} & \cellcolor{fr_purple!8}$\mathbf{62.5}$ & \cellcolor{fr_purple!8}$40.9$ & \cellcolor{fr_purple!8}\underline{$58.1$} & \cellcolor{fr_purple!8}$47.3$
        \\\midrule
        \multirow{8}{*}{\rotatebox[origin=c]{90}{\textbf{nuScenes-C}}} & MinkUNet~\cite{choy2019minkunet} & CVPR'19 & \underline{$100.0$} & $74.4$ & $75.8$ & $53.6$ & $73.9$ & $40.4$ & $\mathbf{73.4}$ & $\mathbf{68.5}$ & $26.6$ & $\mathbf{63.8}$ & $\mathbf{51.0}$
        \\
        & SPVCNN~\cite{tang2020spvnas} & ECCV'20 & $106.7$ & $74.7$ & $74.4$ & $59.0$ & $72.5$ & $41.1$ & $58.4$ & $65.4$ & $36.8$ & $62.3$ & $49.2$
        \\
        & FIDNet~\cite{zhao2021fidnet} & IROS'20 & $122.4$ & $73.3$ & $71.4$ & $64.8$ & $68.0$ & $59.0$ & $48.9$ & $48.1$ & \underline{$57.5$} & $48.8$ & $23.7$
        \\
        & Cylinder3D~\cite{zhu2021cylinder3d} & CVPR'21 & $111.8$ & $72.9$ & $76.2$ & $59.9$ & $72.7$ & $58.1$ & $42.1$ & $64.5$ & $44.4$ & $60.5$ & $42.2$
        \\
        & 2DPASS~\cite{yan20222dpass} & ECCV'22 & $\mathbf{98.6}$ & $75.2$ & $\mathbf{77.9}$ & $64.5$ & $\mathbf{76.8}$ & $54.5$ & \underline{$62.0$} & $67.8$ & $34.4$ & \underline{$63.2$} & \underline{$45.8$}
        \\
        & CENet~\cite{cheng2022cenet} & ICME'22 & $112.8$ & \underline{$76.0$} & $73.3$ & \underline{$67.0$} & $69.9$ & \underline{$61.6$} & $58.3$ & $50.0$ & $\mathbf{60.9}$ & $53.3$ & $24.8$
        \\
        & WaffleIron~\cite{puy2023waffleiron} & ICCV'23 & $106.7$ & $72.8$ & $76.1$ & $56.1$ & $73.9$ & $49.6$ & $59.5$ & $65.2$ & $33.1$ & $61.5$ & $44.0$
        \\
        & \cellcolor{fr_purple!8}\textbf{FRNet} & \cellcolor{fr_purple!8}\textbf{Ours} & \cellcolor{fr_purple!8}$\mathbf{98.6}$ & \cellcolor{fr_purple!8}$\mathbf{77.5}$ & \cellcolor{fr_purple!8}\underline{$77.7$} & \cellcolor{fr_purple!8}$\mathbf{69.1}$ & \cellcolor{fr_purple!8}\underline{$76.6$} & \cellcolor{fr_purple!8}$\mathbf{69.5}$ & \cellcolor{fr_purple!8}$54.5$ & \cellcolor{fr_purple!8}\underline{$68.3$} & \cellcolor{fr_purple!8}$41.4$ & \cellcolor{fr_purple!8}$58.7$ & \cellcolor{fr_purple!8}$43.1$
        \\\bottomrule
    \end{tabular}}
    \label{tab:benchmark_robu}
    \vspace{-0.2cm}
\end{table*}

\noindent\textbf{Label-Efficient LiDAR Semantic Segmentation.}
Recently, semi-supervised learning has gained traction in LiDAR segmentation. In this work, we adopt the approach of LaserMix~\cite{kong2023lasermix} to apply FRNet to the semi-supervised segmentation task, as shown in \cref{tab:benchmark_semi}. For fair comparisons with other methods, FrustumMix and RangeInterpolation are not employed as data augmentation techniques in this benchmark, and TTA is excluded during inference. We first utilize Cylinder3D~\cite{zhu2021cylinder3d} and PTv3~\cite{wu2024ptv3} as backbones and leverage the proposed FrustumMix technique to mix labeled and unlabeled data for consistency learning, which aligns with the LaserMix framework~\cite{kong2023lasermix}. Notably, FrustumMix demonstrates superior performance compared to the original LaserMix~\cite{kong2023lasermix} strategy. Next, we employ FRNet as the backbone and evaluate various strategies for mixing labeled and unlabeled scans, including PolarMix~\cite{xiao2022polarmix}, LaserMix~\cite{kong2023lasermix}, and our FrustumMix. The \textit{Sup.-only} baseline results are also obtained without FrustumMix and RangeInterpolation as data augmentation techniques. The results indicate that FrustumMix consistently achieves the best scores across all data splits when compared with other mixing strategies, particularly with limited annotations.

\noindent\textbf{Out-of-Distribution Robustness.}
Robustness, which reflects a model's ability to generalize under different conditions, is crucial for automotive security systems, particularly in extreme weather conditions~\cite{kong2023robodepth,huang2024pointcat}. To evaluate the out-of-distribution robustness of FRNet, we utilize the noisy datasets SemanticKITTI-C and nuScenes-C introduced in Robo3D~\cite{kong2023robo3d}. In this benchmark, the models trained on SemanticKITTI~\cite{behley2019semantickitti} and nuScenes~\cite{fong2022panoptic} are directly applied to evaluate the performance on SemanticKITTI-C and nuScenes-C~\cite{kong2023robo3d}, respectively, without any fine-tunning. As shown in \cref{tab:benchmark_robu}, FRNet achieves promising performance across most corruption types and demonstrates superiority over recent works employing various LiDAR representations, including sparse voxel~\cite{choy2019minkunet,tang2020spvnas,zhu2021cylinder3d}, range view~\cite{zhao2021fidnet,cheng2022cenet}, multi-view fusion~\cite{yan20222dpass}, and raw points~\cite{puy2023waffleiron}.

\begin{figure*}[t]
    \centering
    \includegraphics[width=1.0\linewidth]{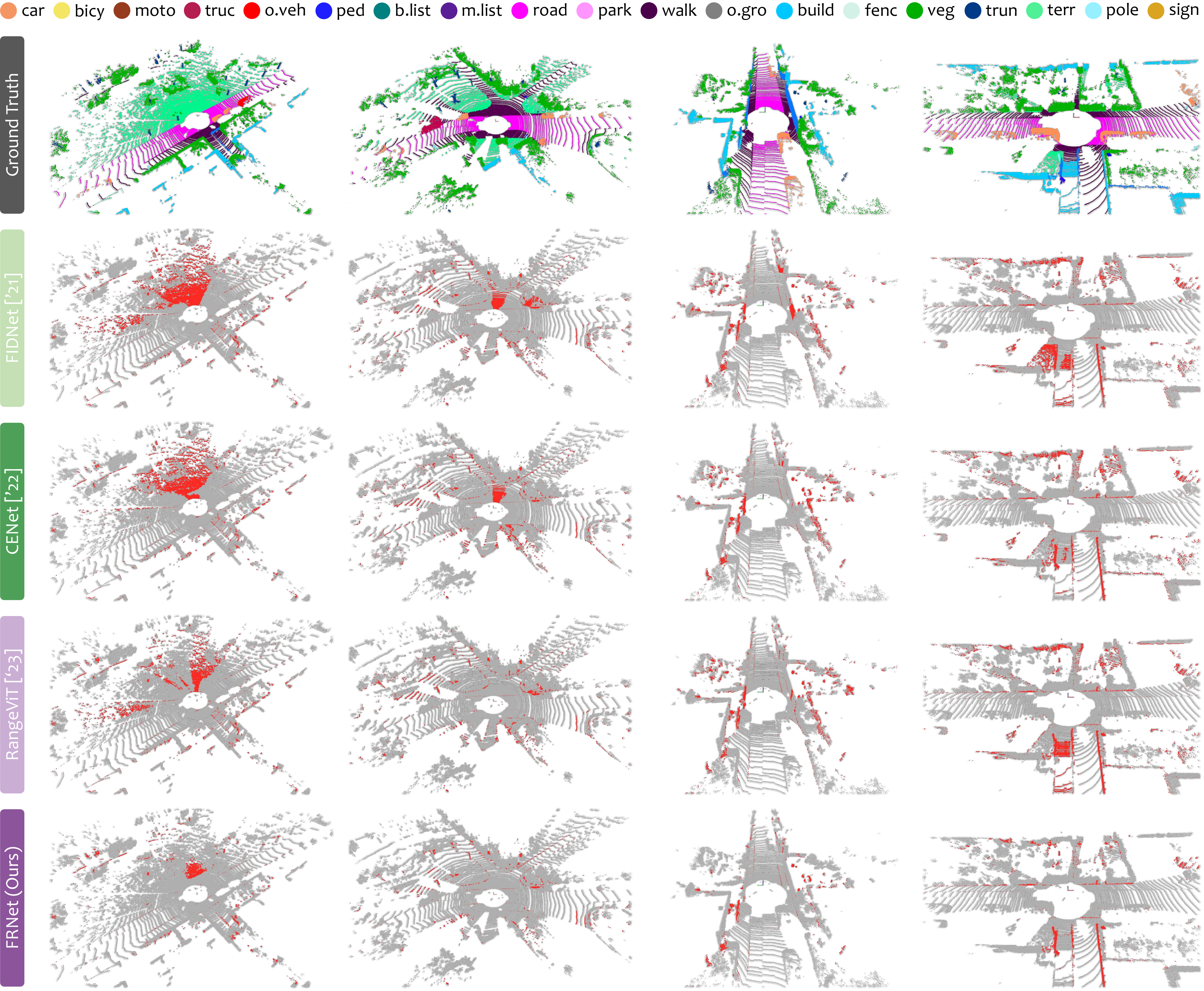}
    \caption{\textbf{Qualitative results} among state-of-the-art LiDAR segmentation methods~\cite{zhao2021fidnet,cheng2022cenet,ando2023rangevit} on the \textit{val} set of SemanticKITTI~\cite{behley2019semantickitti}. To highlight the differences compared with groundtruth, the \textbf{\textcolor[RGB]{206,206,206}{correct}} and \textbf{\textcolor[RGB]{192,0,0}{incorrect}} predictions are painted in \textbf{\textcolor[RGB]{206,206,206}{gray}} and \textbf{\textcolor[RGB]{192,0,0}{red}}, respectively. Best viewed in colors and zoomed-in for details.}
    \label{fig:vis}
    \vspace{-0.4cm}
\end{figure*}

\noindent\textbf{Quantitative Assessments.}
We compare the class-wise IoU scores of FRNet with those of the baseline method~\cite{cheng2022cenet} in \cref{fig:classwise}. Notably, FRNet demonstrates significant improvement across most semantic classes, particularly for dynamic classes with intricate structures, where it achieves mIoU gains ranging from $15\%$ to $24\%$. In \cref{fig:vis}, we present the prediction results on SemanticKITTI~\cite{behley2019semantickitti}, compared with those of state-of-the-art range view methods, including FIDNet~\cite{zhao2021fidnet}, CENet~\cite{cheng2022cenet}, and RangeViT~\cite{ando2023rangevit}. It is evident that FRNet delivers more accurate predictions, particularly on large planar objects such as ``vegetation'', ``terrain'', and ``sidewalk'', which are often challenging to distinguish accurately.

\subsection{Ablation Study}

In this section, we will discuss the efficacy of each design in our FRNet architecture. Unless otherwise specified, all experiments are conducted and reported on the \textit{val} set of SemanticKITTI~\cite{behley2019semantickitti} and nuScenes~\cite{fong2022panoptic}, respectively.

\begin{table}[t]
    \centering
    \caption{\textbf{Ablation study of each component in FRNet} on the \textit{val} set of SemanticKITTI~\cite{behley2019semantickitti} and nuScenes~\cite{fong2022panoptic}. \textbf{FFE}: Frustum Feature Encoder. \textbf{FP}: Frustum-Point Fusion. \textbf{FH}: Fusion Head. \textbf{FS}: Frustum-level Supervision. \textbf{RI}: RangeInterpolation. \textbf{TTA}: Test Time Augmentation. All \textbf{mIoU} and \textbf{mAcc} scores are given in percentage (\%).}
    \resizebox{\linewidth}{!}{
    \begin{tabular}{cccccc|p{14pt}<{\centering}p{14pt}<{\centering}|p{14pt}<{\centering}p{14pt}<{\centering}}
        \toprule
        \multirow{2}{*}{\textbf{FFE}} & \multirow{2}{*}{\textbf{FP}} & \multirow{2}{*}{\textbf{FH}} & \multirow{2}{*}{\textbf{FS}} & \multirow{2}{*}{\textbf{RI}} & \multirow{2}{*}{\textbf{TTA}} & \multicolumn{2}{c|}{\textbf{SemKITTI}} & \multicolumn{2}{c}{\textbf{nuScenes}}
        \\
        & & & & & & \textbf{mIoU} & \textbf{mAcc} & \textbf{mIoU} & \textbf{mAcc}
        \\\midrule\midrule
        \cellcolor{fr_green!8}\checkmark & \cellcolor{fr_green!8} & \cellcolor{fr_green!8} & \cellcolor{fr_green!8} & \cellcolor{fr_green!8} & \cellcolor{fr_green!8} & \cellcolor{fr_green!8}$62.7$ & \cellcolor{fr_green!8}$72.3$ & \cellcolor{fr_green!8}$72.8$ & \cellcolor{fr_green!8}$82.7$
        \\\midrule
        \checkmark & \checkmark & & & & & $64.3$ & $73.5$ & $73.6$ & $83.4$
        \\
        \checkmark & \checkmark & & \checkmark & & & $66.2$ & $73.9$ & $76.4$ & $84.6$
        \\
        \checkmark & \checkmark & \checkmark & \checkmark & & & $67.0$ & $74.1$ & $77.2$ & $84.8$
        \\
        \checkmark & \checkmark & \checkmark & \checkmark & \checkmark & & $67.6$ & $74.2$ & $77.7$ & $\mathbf{85.2}$
        \\\midrule
        \cellcolor{fr_purple!8}\checkmark & \cellcolor{fr_purple!8}\checkmark & \cellcolor{fr_purple!8}\checkmark & \cellcolor{fr_purple!8}\checkmark & \cellcolor{fr_purple!8}\checkmark & \cellcolor{fr_purple!8}\checkmark & \cellcolor{fr_purple!8}$\mathbf{68.7}$ & \cellcolor{fr_purple!8}$\mathbf{74.9}$ & \cellcolor{fr_purple!8}$\mathbf{79.0}$ & \cellcolor{fr_purple!8}$85.0$
        \\\bottomrule
    \end{tabular}}
    \label{tab:ablation}
    \vspace{-0.3cm}
\end{table}

\noindent\textbf{Component Designs.}
\cref{tab:ablation} summarizes the ablation results of each component in the FRNet architecture. Notably, FrustumMix is employed across all setups in this ablation study. Firstly, we directly apply the frustum feature encoder to CENet~\cite{cheng2022cenet}, incorporating KNN post-processing to predict semantic labels over the entire point cloud, achieving $62.7\%$ and $72.8\%$ mIoU on SemanticKITTI~\cite{behley2019semantickitti} and nuScenes~\cite{fong2022panoptic}, respectively. Next, by integrating point-frustum fusion to hierarchically update point features, the semantic results are predicted with the updated point features in an end-to-end manner, resulting in improvements of $1.6\%$ and $0.8\%$ mIoU, respectively. To regularize the frustum features, the frustum-level supervision is introduced and it leads to significant gains of $1.9\%$ and $2.8\%$ mIoU, respectively. The fusion head module, which fuses multiple features at different levels for more accurate prediction, brings an additional $0.8\%$ mIoU gain on both datasets. Furthermore, the proposed RangeInterpolation, aimed at alleviating empty pixels in the 2D representation by reconstructing semantic surfaces via surrounding range information, provides a performance boost of approximately $0.5\%$ mIoU. Finally, adopting Test Time Augmentation during inference, as demonstrated in prior works, leads to improvements of $1.1\%$ and $1.3\%$ mIoU, respectively.

\begin{table}[t]
    \centering
    \caption{\textbf{Ablation study of frustum resolution in FRNet} on the \textit{val} set of SemanticKITTI~\cite{behley2019semantickitti}. All \textbf{mIoU} and \textbf{mAcc} scores are given in percentage ($\%$) without TTA.}
    \label{tab:resolution}
    \resizebox{\linewidth}{!}{
    \begin{tabular}{c|cc|cc}
        \toprule
        \textbf{Number of Frustums} & \textbf{Height} $(H)$ & \textbf{Width} $(W)$ & \textbf{mIoU} & \textbf{mAcc}
        \\\midrule\midrule
        $131,072$ & $64$ & $2048$ & $66.8$ & $73.8$
        \\
        $65,536$ & $64$ & $1024$ & $67.1$ & $73.4$
        \\
        $65,536$ & $32$ & $2048$ & $65.5$ & $72.7$
        \\
        \cellcolor{fr_purple!8}$32,768$ & \cellcolor{fr_purple!8}$64$ & \cellcolor{fr_purple!8}$512$ & \cellcolor{fr_purple!8}$\mathbf{67.6}$ & \cellcolor{fr_purple!8}$\mathbf{74.2}$
        \\
        $32,768$ & $32$ & $1024$ & $66.0$ & $72.4$
        \\
        $16,384$ & $64$ & $256$ & $64.3$ & $70.5$
        \\
        $16,384$ & $32$ & $512$ & $65.5$ & $72.1$
        \\
        $8,192$ & $32$ & $256$ & $64.3$ & $70.8$
        \\\bottomrule
    \end{tabular}}
    \vspace{-0.3cm}
\end{table}

\noindent\textbf{Frustum Representation Resolution.}
Finding a suitable frustum resolution is crucial for balancing accuracy and efficiency. In \cref{tab:resolution}, we conduct a series of experiments with various frustum resolutions. While higher resolution can capture more detailed features, it may not always lead to optimal results. The limited area of the frustum region can result in a lack of meaningful structure, causing the model to struggle with noisy or sparse data. Conversely, although lower resolution can improve inference efficiency and provide a more coherent representation, the larger area of the frustum region may introduce excessive noise information, particularly at object boundaries. Thus, finding an optimal frustum resolution is crucial for balancing the richness of features with computational demands and maintaining overall model performance.

\begin{figure}
    \centering
    \includegraphics[width=1.0\linewidth]{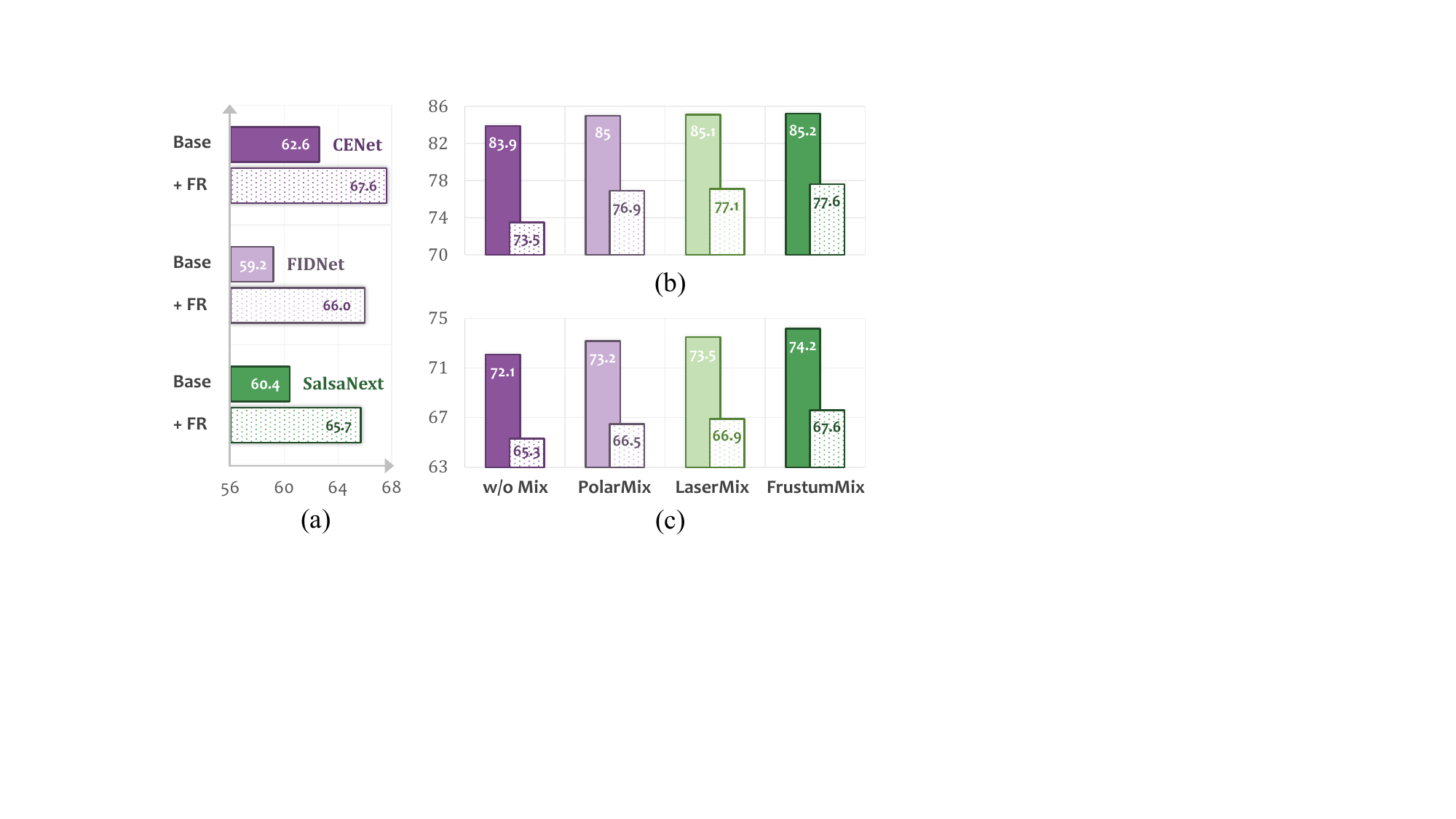}
    \caption{\textbf{Ablation study} on different FRNet configurations. Plots show the comparisons of: a) Different range-view backbones on the \textit{val} set of SemanticKITTI~\cite{behley2019semantickitti}. b) Different mixing strategies on the \textit{val} set of nuScenes~\cite{fong2022panoptic}. c) Different mixing strategies on the \textit{val} set of SemanticKITTI~\cite{behley2019semantickitti}. Color-filled boxes denote \textbf{mAcc} while empty boxes denote \textbf{mIoU} in (b) and (c), respectively.}
    \label{fig:compare}
\end{figure}

\noindent\textbf{Mixing Strategies.}
We further explore the effectiveness of the proposed FrustumMix. As shown in \cref{tab:benchmark_semi}, FrustumMix significantly improves performance on semi-supervised learning tasks and demonstrates generalizability to other modalities. To contextualize FrustumMix within common data augmentation techniques, we compare it with recent popular outdoor mixing strategies, including LaserMix~\cite{kong2023lasermix} and PolarMix~\cite{xiao2022polarmix}. As illustrated in \cref{fig:compare}(b) and \cref{fig:compare}(c), FrustumMix consistently outperforms these prior strategies.

\begin{table}[t]
    \centering
    \caption{\textbf{Ablation study on window settings} in RangeInterpolation on the \textit{val} set of SemanticKITTI~\cite{behley2019semantickitti}. \textcolor{green}{Green} pixel denotes the empty pixel needed to be interpolated. \textcolor{violet}{Violet} pixels are the surrounding pixels used to generate the point. All \textbf{mIoU} scores are given in percentage (\%).}
    \resizebox{\linewidth}{!}{
    \begin{tabular}{p{40pt}<{\centering}|p{40pt}<{\centering}|p{40pt}<{\centering}|p{40pt}<{\centering}|p{40pt}<{\centering}}
        \toprule
        \cellcolor{fr_green!8}Baseline & \cellcolor{fr_green!8}$1\times3$ & \cellcolor{fr_green!8}$3\times1$ & \cellcolor{fr_green!8}$1\times5$ & \cellcolor{fr_green!8}$5\times1$
        \\\midrule
        \begin{minipage}[b]{0.16\columnwidth}\centering\raisebox{-.4\height}{\includegraphics[width=\linewidth]{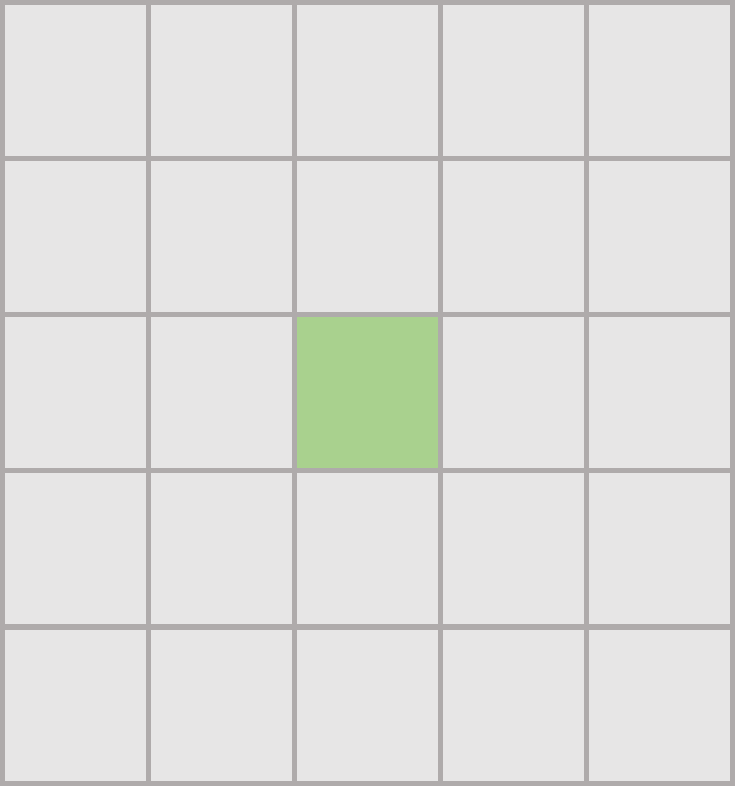}}\end{minipage} & \begin{minipage}[b]{0.16\columnwidth}\centering\raisebox{-.4\height}{\includegraphics[width=\linewidth]{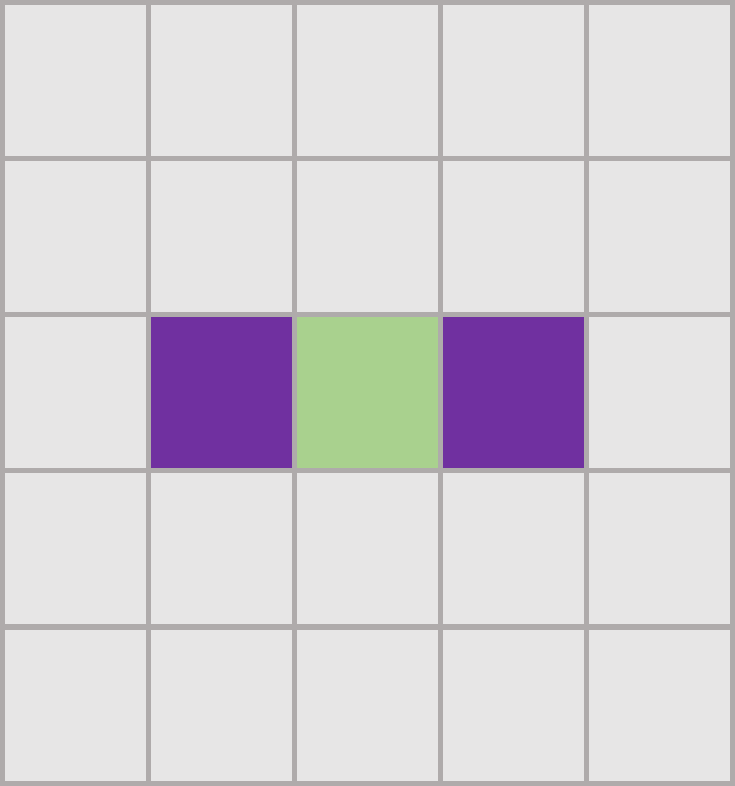}}\end{minipage} & \begin{minipage}[b]{0.16\columnwidth}\centering\raisebox{-.4\height}{\includegraphics[width=\linewidth]{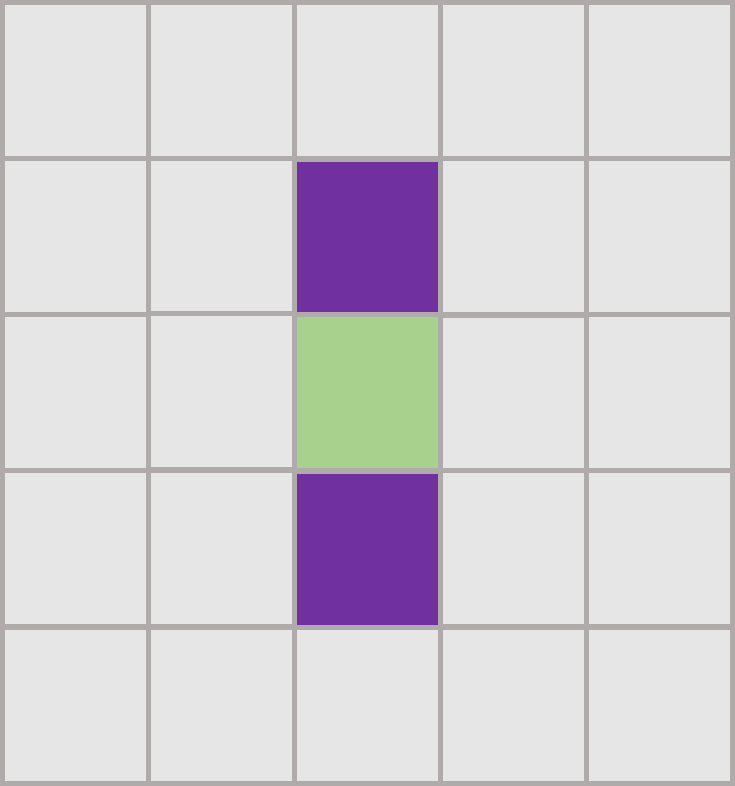}}\end{minipage} & \begin{minipage}[b]{0.16\columnwidth}\centering\raisebox{-.4\height}{\includegraphics[width=\linewidth]{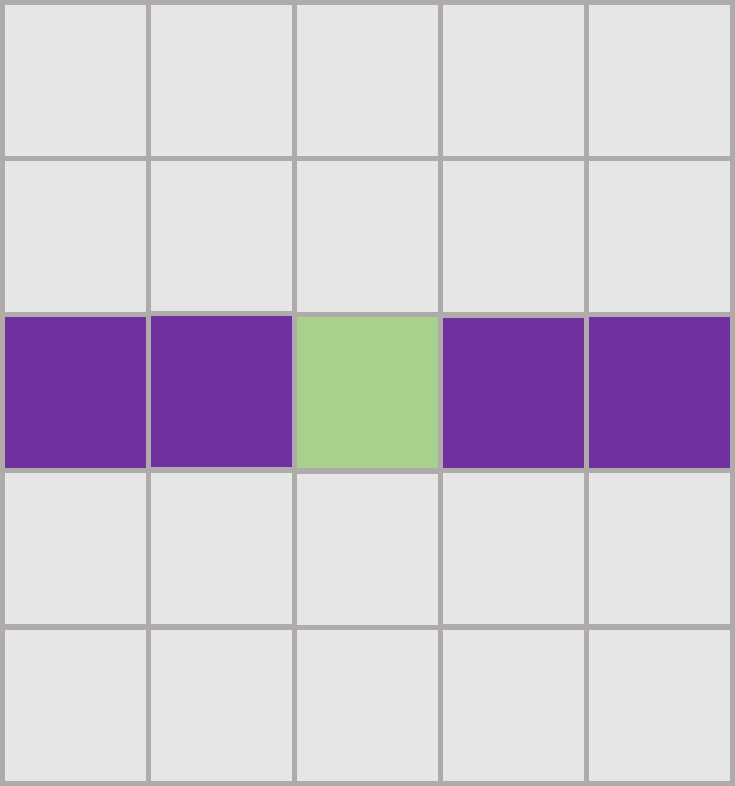}}\end{minipage} & \begin{minipage}[b]{0.16\columnwidth}\centering\raisebox{-.4\height}{\includegraphics[width=\linewidth]{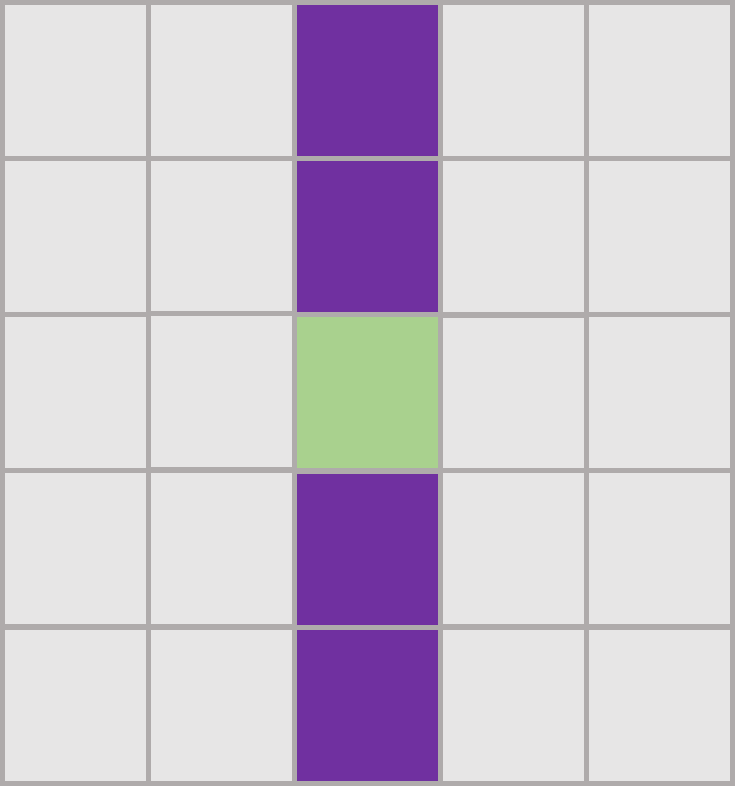}}\end{minipage}
        \\\midrule
        $67.0$ & $67.6_\mathbf{{\textcolor{fr_green}{(+0.6)}}}$ & $67.4_\mathbf{{\textcolor{fr_green}{(+0.4)}}}$ & $67.5_\mathbf{{\textcolor{fr_green}{(+0.5)}}}$ & $67.4_\mathbf{{\textcolor{fr_green}{(+0.4)}}}$
        \\\midrule
        \cellcolor{fr_green!8}cross $3\times3$ & \cellcolor{fr_green!8}cross $5\times5$ & \cellcolor{fr_green!8}$3 \times 5$ & \cellcolor{fr_green!8}$3\times3$ & \cellcolor{fr_green!8}$5\times5$
        \\\midrule
        \begin{minipage}[b]{0.16\columnwidth}\centering\raisebox{-.4\height}{\includegraphics[width=\linewidth]{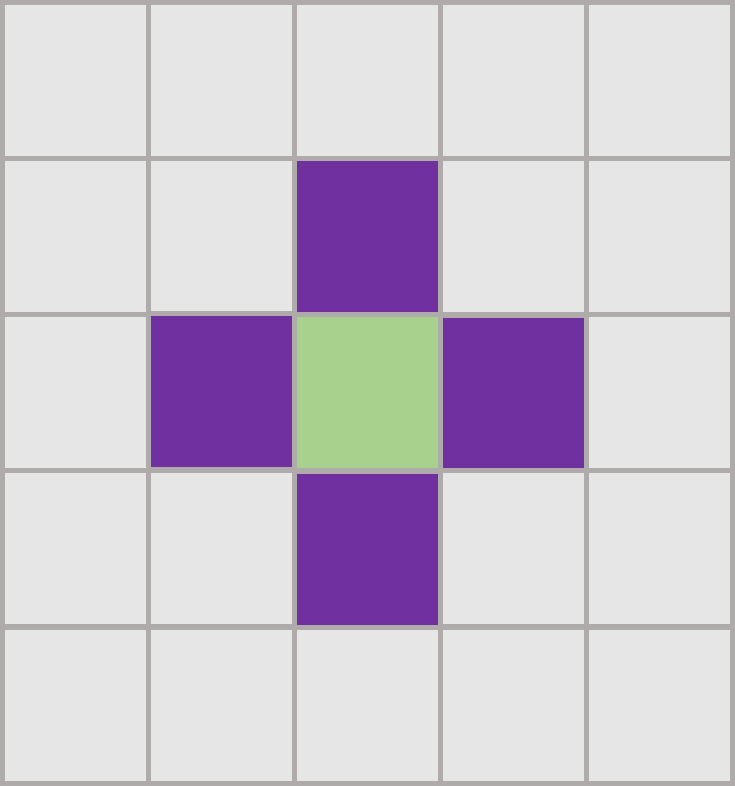}}\end{minipage} & \begin{minipage}[b]{0.16\columnwidth}\centering\raisebox{-.4\height}{\includegraphics[width=\linewidth]{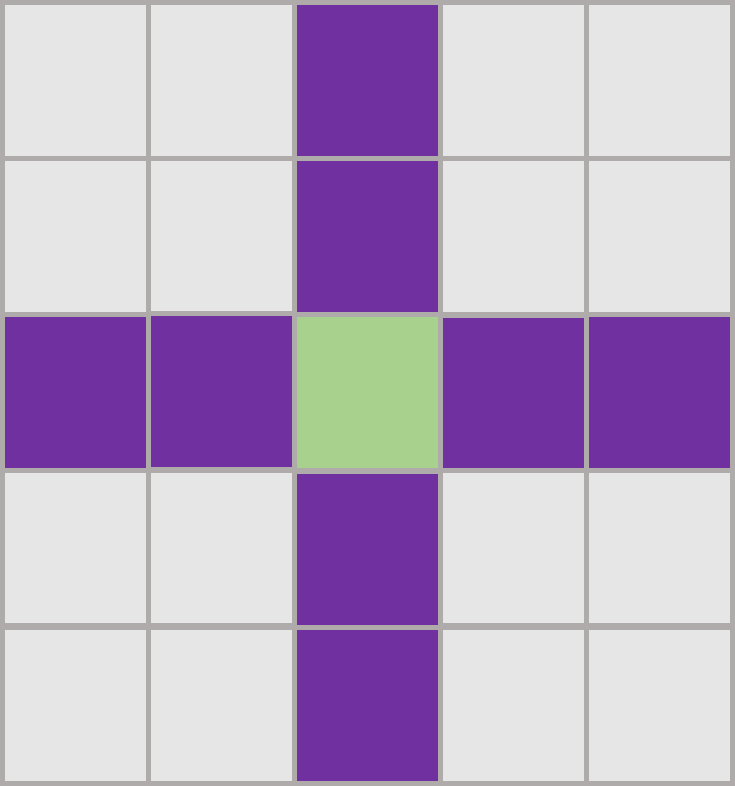}}\end{minipage} & \begin{minipage}[b]{0.16\columnwidth}\centering\raisebox{-.4\height}{\includegraphics[width=\linewidth]{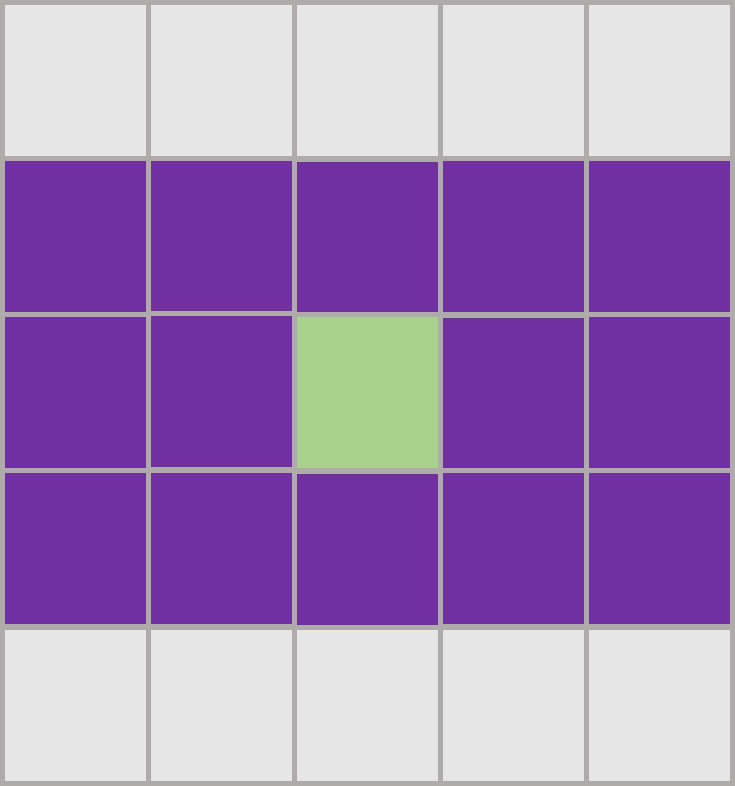}}\end{minipage} & \begin{minipage}[b]{0.16\columnwidth}\centering\raisebox{-.4\height}{\includegraphics[width=\linewidth]{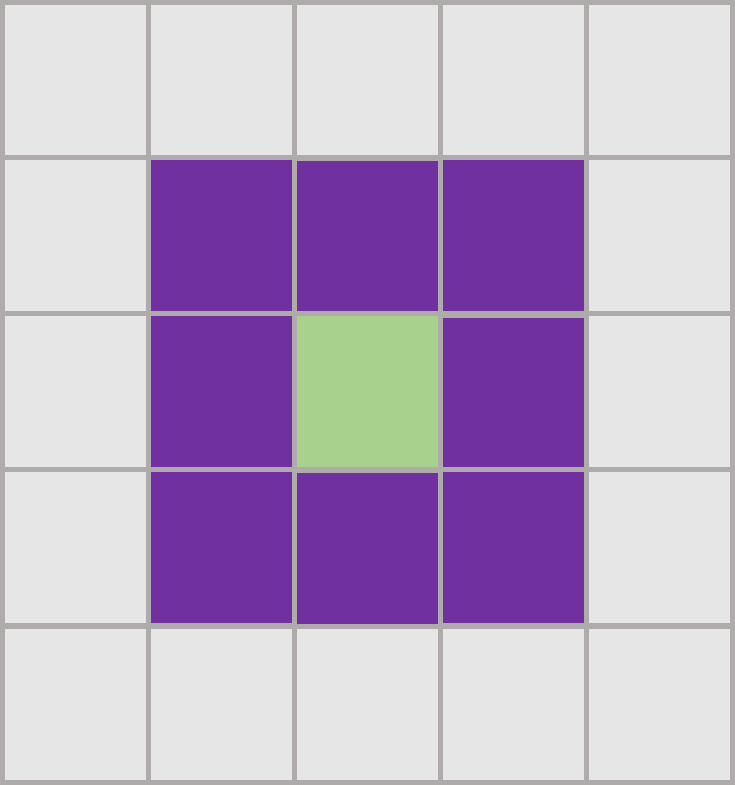}}\end{minipage} & \begin{minipage}[b]{0.16\columnwidth}\centering\raisebox{-.4\height}{\includegraphics[width=\linewidth]{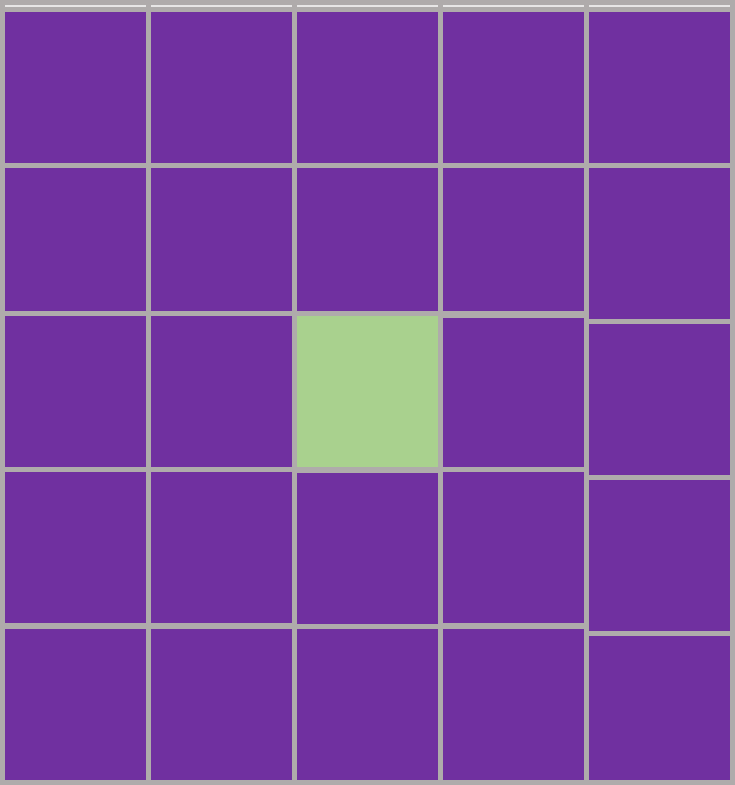}}\end{minipage}
        \\\midrule
        $67.5_\mathbf{{\textcolor{fr_green}{(+0.5)}}}$ & $67.4_\mathbf{{\textcolor{fr_green}{(+0.4)}}}$ & $67.1_\mathbf{{\textcolor{fr_green}{(+0.1)}}}$ & $66.8_\mathbf{{\textcolor{fr_purple}{(-0.2)}}}$ & $66.6_\mathbf{{\textcolor{fr_purple}{(-0.4)}}}$
        \\\bottomrule
    \end{tabular}}
    \label{tab:range_interpolation}
    \vspace{-0.4cm}
\end{table}

\noindent\textbf{RangeInterpolation Window Settings.}
RangeInterpolation reconstructs semantic surfaces based on the range image, facilitating a compact 2D representation for coherent semantic learning. \cref{tab:range_interpolation} showcases the performance of RangeInterpolation with different window settings. Notably, we observe consistent performance improvements when considering interpolation along the horizontal or vertical direction exclusively. Specifically, configurations employing horizontal windows, such as the $1\times3$ and $1\times5$ settings, demonstrate superior results compared to those focusing on the vertical direction ($3\times1$ and $5\times1$ windows). We conjecture that points aligned along the same laser beams often share similar attributes, leading to more accurate and coherent semantic surfaces. Conversely, larger window sizes, such as the $3\times5$, $3\times3$, and $5\times5$ configurations, tend to exhibit non-uniform point distributions within the window, resulting in noisy points that fail to accurately represent the expected region and consequently leading to a drop in performance.

\noindent\textbf{2D Backbones.}
To validate the versatility of our method, we employ various range-view methods as our 2D backbone for extracting frustum features, including SalsaNext~\cite{cortinhal2020salsanext}, FIDNet~\cite{zhao2021fidnet}, and CENet~\cite{cheng2022cenet}. As depicted in \cref{fig:compare}(a), our proposed frustum-range representation consistently delivers performance gains ranging from $5.0\%$ to $6.8\%$ in mIoU across different range-view methods.

\begin{table}[t]
    \centering
    \caption{\textbf{Inference efficiency comparison} of FRNet and MinkUNet~\cite{choy2019minkunet} implemented with various codebases and sparse convolution backends, as well as PTv3~\cite{wu2024ptv3} on the \textit{val} set of SemanticKITTI~\cite{behley2019semantickitti}.}
    \label{tab:inference}
    \begin{tabular}{c|c|c|c}
        \toprule
        \textbf{Codebase} & \textbf{Backbone} & \textbf{Backend} & \textbf{FPS}
        \\\midrule\midrule
        \multirow{6}{*}{MMDetection3D} & MinkUNet & MinkowskiEngine & $9.1$
        \\
        & MinkUNet & SpConv & $8.6$
        \\
        & MinkUNet & TorchSparse & $10.2$
        \\
        & MinkUNet & TorchSparse++ & $21.3$
        \\
        & \cellcolor{fr_purple!8}FRNet & \cellcolor{fr_purple!8}Convolution & \cellcolor{fr_purple!8}$29.1$
        \\
        & \cellcolor{fr_purple!8}Fast-FRNet & \cellcolor{fr_purple!8}Convolution & \cellcolor{fr_purple!8}$\mathbf{33.8}$
        \\\midrule
        PCSeg & MinkUNet & TorchSparse & $10.6$
        \\\midrule
        \multirow{2}{*}{PointCept} & MinkUNet & MinkowskiEngine & $8.9$
        \\
        & PTv3 & FlashAttention & $20.2$
        \\\bottomrule
    \end{tabular}
\end{table}

\noindent\textbf{Inference Efficiency.}
Multiple sparse convolution backends, including MinknowskiEngine, SpConv, TorchSparse, and TorchSparse++, offer efficient processing for voxel-based methods. To compare FRNet with these backends, we implement MinkUNet~\cite{choy2019minkunet} using the MMDetection3D codebase across various backends. Additionally, we evaluate the FPS of MinkUNet implemented with the PCSeg and PointCept codebases, as well as the recent efficient method PTv3~\cite{wu2024ptv3}. All evaluations are conducted on a single NVIDIA 2080Ti GPU. As shown in \cref{tab:inference}, while TorchSparse++ delivers satisfactory efficiency for LiDAR segmentation, FRNet achieves the fastest inference speed, making it suitable for real-time applications in autonomous driving systems.

\begin{figure*}[t]
    \centering
    \begin{subfigure}[b]{0.31\linewidth}
        \centering
        \includegraphics[width=\linewidth]{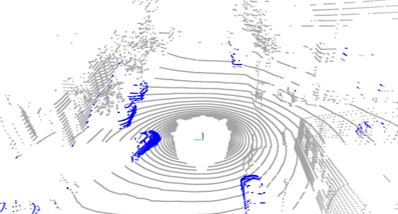}
        \caption{car}
        \label{fig_rebuttal:car}
    \end{subfigure}
    ~~
    \begin{subfigure}[b]{0.31\linewidth}
        \centering
        \includegraphics[width=\linewidth]{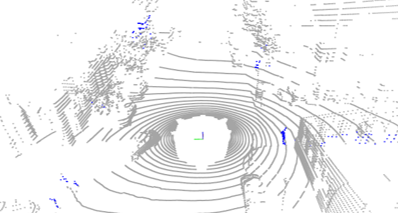}
        \caption{person}
        \label{fig_rebuttal:person}
    \end{subfigure}
    ~~
    \begin{subfigure}[b]{0.31\linewidth}
        \centering
        \includegraphics[width=\linewidth]{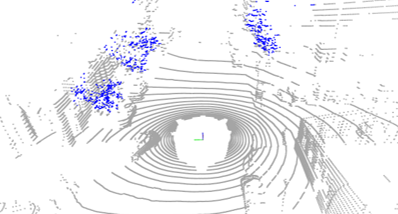}
        \caption{vegetation}
        \label{fig_rebuttal:vegetation}
    \end{subfigure}
    \begin{subfigure}[b]{0.31\linewidth}
        \centering
        \includegraphics[width=\linewidth]{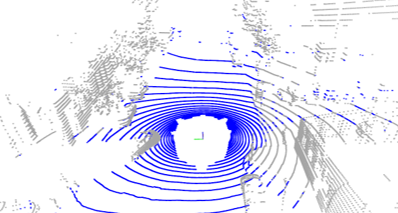}
        \caption{drivable surface}
        \label{fig_rebuttal:drivable}
    \end{subfigure}
    ~~
    \begin{subfigure}[b]{0.31\linewidth}
        \centering
        \includegraphics[width=\linewidth]{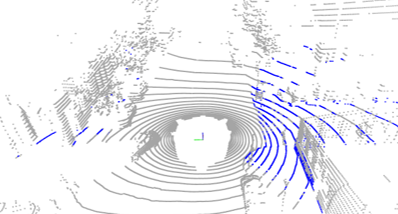}
        \caption{sidewalk}
        \label{fig_rebuttal:sidewalk}
    \end{subfigure}
    ~~
    \begin{subfigure}[b]{0.31\linewidth}
        \centering
        \includegraphics[width=\linewidth]{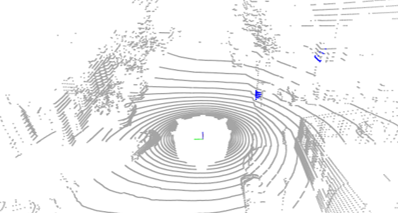}
        \caption{trailer}
        \label{fig_rebuttal:trailer}
    \end{subfigure}
    \caption{\textbf{The open-vocabulary LiDAR semantic segmentation experiments of FRNet.} Our model demonstrates the ability to perform open-vocabulary segmentation with various textual inputs. The segmented points corresponding to the provided text descriptions are highlighted in \textcolor{blue}{blue}, showcasing its potential for open-world perception tasks.}
    \label{fig:open_vocabulary}
    \vspace{-0.4cm}
\end{figure*}

\begin{table}[t]
    \centering
    \caption{\textbf{The multi-source training results of the proposed Fast-FRNet} on the \textit{val} set of the SemanticKITTI~\cite{behley2019semantickitti}, nuScenes~\cite{fong2022panoptic}, and SemanticPOSS~\cite{pan2020semanticposs} datasets. All \textbf{mIoU} and \textbf{mAcc} scores are given in percentage (\%).}
    \resizebox{1.0\linewidth}{!}{
    \begin{tabular}{c|cc|cc|cc}
        \toprule
        \multirow{2}{*}{\textbf{Method}} & \multicolumn{2}{c|}{\textbf{SemKITTI}} & \multicolumn{2}{c|}{\textbf{nuScenes}} & \multicolumn{2}{c}{\textbf{SemPOSS}}
        \\
        & \textbf{mIoU} & \textbf{mAcc} & \textbf{mIoU} & \textbf{mAcc} & \textbf{mIoU} & \textbf{mAcc}
        \\\midrule\midrule
        Single-Source & $67.1$ & $73.8$ & $78.8$ & $84.4$ & $52.5$ & $67.1$
        \\
        Multi-Source & $67.4$ & $74.2$ & $79.1$ & $84.7$ & $53.2$ & $68.0$
        \\
        \textbf{Fast-FRNet} (w/ PPT~\cite{wu2024ppt}) & $\mathbf{68.2}$ & $\mathbf{74.7}$ & $\mathbf{79.3}$ & $\mathbf{84.9}$ & $\mathbf{53.9}$ & $\mathbf{68.4}$
        \\\bottomrule
    \end{tabular}}
    \label{tab:multi_dataset}
    \vspace{-0.4cm}
\end{table}

\begin{table}[t]
    \centering
    \caption{\textbf{Zero-shot segmentation performance} on the \textit{val} set of nuScenes~\cite{fong2022panoptic}. All \textbf{mIoU} and \textbf{mAcc} scores are given in percentage (\%).}
    \begin{tabular}{c|cc}
        \toprule
        \multirow{2}{*}{\textbf{Method}} & \multicolumn{2}{c}{\textbf{nuScenes}}
        \\
        & \textbf{mIoU} & \textbf{mAcc}
        \\\midrule\midrule
        OpenScene~\cite{peng2023openscene} & $42.1$ & $61.8$
        \\
        + \textbf{FRNet} & $\mathbf{43.5}$ & $\mathbf{62.0}$
        \\\bottomrule
    \end{tabular}
    \label{tab:zeroshot}
    \vspace{-0.4cm}
\end{table}

\noindent\textbf{Multi-Dataset Fusion.}
Recent studies~\cite{wu2024ppt,liu2024m3net} highlight the importance of large-scale benchmarks achieved through multi-dataset fusion strategies across heterogeneous datasets. In this work, We extend Fast-FRNet to support multi-dataset joint training, as shown in \cref{tab:multi_dataset}. Initially, we combine multi-source datasets using mean-variance normalization to align intensity values, which provides slight performance gains compared to single-source training. Additionally, inspired by PPT~\cite{wu2024ppt}, we incorporate point-prompt techniques to facilitate multi-source joint training. This approach yields further improvements, demonstrating the effectiveness of prompt-guided strategies in multi-dataset fusion.

\noindent\textbf{Open-Vocabulary Semantic Segmentation.}
To investigate the potential of FRNet in open-vocabulary segmentation, we adopt an approach inspired by OpenScene~\cite{peng2023openscene}. Specifically, we integrate text and image features into the framework, aligning them with point features. During inference, text inputs are used to guide the segmentation of point clouds with similar features. As shown in \cref{tab:zeroshot}, FRNet demonstrates promising performance in zero-shot segmentation. Additionally, in \cref{fig:open_vocabulary}, we present qualitative results of segmentation using corresponding text inputs, highlighting the capability of FRNet to handle open-world perception tasks effectively.

\section{Conclusion}
\label{sec:conclusion}

In this work, we presented FRNet, a Frustum-Range network designed for efficient and effective LiDAR segmentation, which can be seamlessly integrated into range-view approaches. FRNet comprises three key components: the Frustum Feature Encoder Module for extracting per-point features at lower levels, the Frustum-Point Fusion Module for hierarchical updating of per-point features, and the Fusion Head Module for fusing features from different levels to predict more accurate labels. Additionally, we introduced two efficient augmentation methods, FurstumMix and RangeInterpolation, to enhance the generalizability of range-view LiDAR segmentation. Extensive experiments across diverse driving perception datasets verified that FRNet achieves competitive performance while maintaining high efficiency. We hope this work can lay a solid foundation for future works targeted at real-time, in-vehicle 3D perception. Moving forward, we will further enhance the representation learning capability of FRNet and incorporate such a method into other 3D perception tasks, such as 3D object detection and semantic occupancy prediction, and comprehensive vehicle-road coordination perception tasks.

\section*{Appendix}

\section{Additional Experimental Results}
\label{sec:additional-results}

In this section, we provide the complete experimental results among the four popular LiDAR segmentation benchmarks used in this work.

\begin{figure*}[t]
    \vspace{-0.37cm}
    \centering
    \begin{subfigure}[b]{.48\textwidth}
        \centering
        \includegraphics[width=\textwidth]{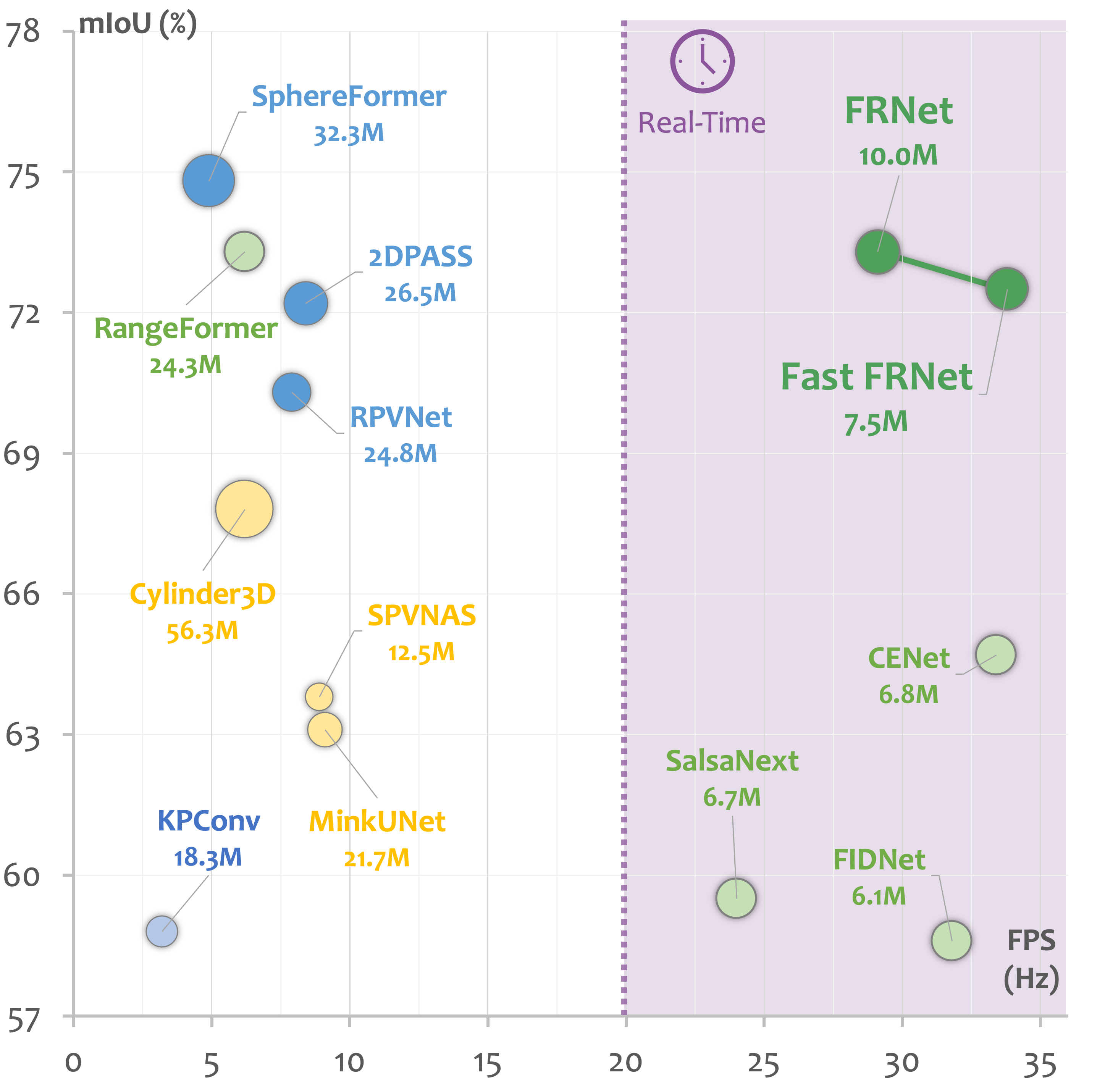}
        \caption{Speed \textit{vs.} Accuracy}
        \label{fig:scalability_miou}
    \end{subfigure}
    \hfill
    \begin{subfigure}[b]{.48\textwidth}
         \centering
         \includegraphics[width=\textwidth]{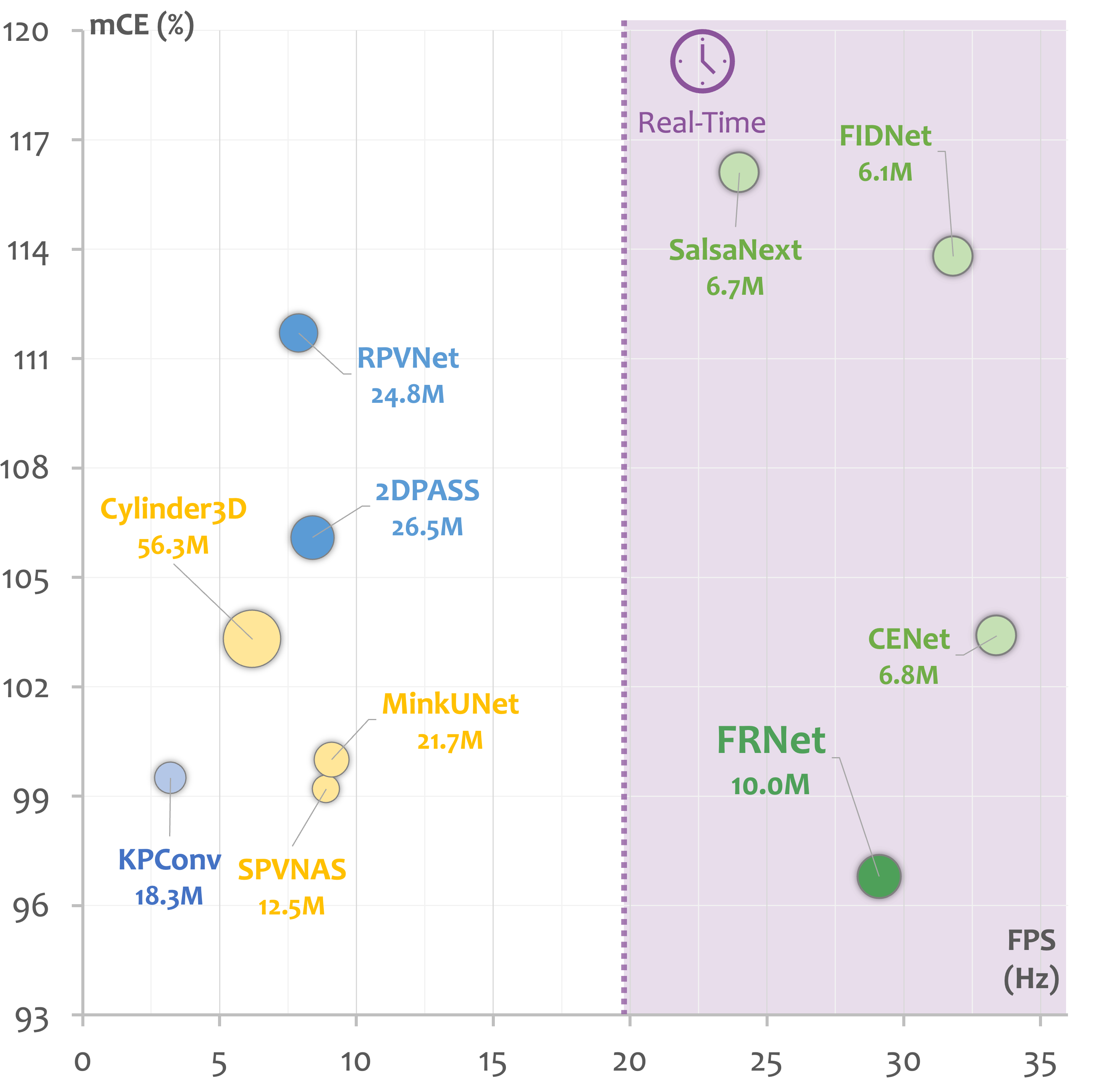}
         \caption{Speed \textit{vs.} Robustness}
         \label{fig:scalability_mce}
    \end{subfigure}
    \caption{\textbf{The scalability analysis} of existing LiDAR semantic segmentation approaches. Subfigure (a) The inference speed \textit{vs.} segmentation accuracy (the higher the better). Subfigure (b) The inference speed \textit{vs.} corruption error (the lower the better).}
    \label{fig:scalability}
\end{figure*}

\subsection{Scalability and Robustness}
In this section, we provide additional analysis of existing LiDAR segmentation approaches regarding their trade-offs in-between the inference speed (FPS), segmentation accuracy (mIoU), and out-of-training-distribution robustness (mCE). As shown in \cref{fig:scalability}, current LiDAR segmentation models pursue either the segmentation accuracy or inference speed. Those more accurate LiDAR segmentation models often contain larger parameter sets and are less efficient during the inference stage. This is particularly predominant for the voxel-based (as shown in yellow circulars) and point-voxel fusion (as shown in blue circulars) approaches. The range-view approaches, on the contrary, are faster in terms of inference speed. However, existing range-view models only achieve sub-par performance compared to the voxel-based and point-voxel fusion approaches. It is worth highlighting again that the proposed FRNet and Fast-FRNet provide a better trade-off in-between speed, accuracy, and robustness. Our models achieve competitive segmentation results and superior robustness over the voxel-based and point-voxel fusion counterparts, while still maintaining high efficiency for real-time LiDAR segmentation.

\begin{table*}[t]
\caption{\textbf{The class-wise IoU scores} of different LiDAR semantic segmentation approaches on the SemanticKITTI~\cite{behley2019semantickitti} leaderboard. All IoU scores are given in percentage (\%). The \textbf{best} and \underline{second best} scores for each class are highlighted in \textbf{bold} and \underline{underline}.}
\vspace{-0.1cm}
\begin{adjustbox}{width=\textwidth}
\begin{tabular}{r|c|ccccccccccccccccccc}
\toprule
\textbf{Method} & \rotatebox{90}{$\text{mIoU}$} & \rotatebox{90}{car} & \rotatebox{90}{bicycle} & \rotatebox{90}{motorcycle} & \rotatebox{90}{truck} & \rotatebox{90}{other-vehicle~} & \rotatebox{90}{person} & \rotatebox{90}{bicyclist} & \rotatebox{90}{motorcyclist} & \rotatebox{90}{road} & \rotatebox{90}{parking} & \rotatebox{90}{sidewalk} & \rotatebox{90}{other-ground~} & \rotatebox{90}{building} & \rotatebox{90}{fence} & \rotatebox{90}{vegetation} & \rotatebox{90}{trunk} & \rotatebox{90}{terrain} & \rotatebox{90}{pole} & \rotatebox{90}{traffic-sign}
\\\midrule\midrule
PointNet~\cite{qi2017pointnet} & $14.6$ & $46.3$ & $1.3$ & $0.3$ & $0.1$ & $0.8$ & $0.2$ & $0.2$ & $0.0$ & $61.6$ & $15.8$ & $35.7$ & $1.4$ & $41.4$ & $12.9$ & $31.0$ & $4.6$ & $17.6$ & $2.4$ & $3.7$
\\
PointNet++~\cite{qi2017pointnet++} & $20.1$ & $53.7$ & $1.9$ & $0.2$ & $0.9$ & $0.2$ & $0.9$ & $1.0$ & $0.0$ & $72.0$ & $18.7$ & $41.8$ & $5.6$ & $62.3$ & $16.9$ & $46.5$ & $13.8$ & $30.0$ & $6.0$ & $8.9$
\\
SqSeg~\cite{wu2018squeezeseg} & $30.8$ & $68.3$ & $18.1$ & $5.1$ & $4.1$ & $4.8$ & $16.5$ & $17.3$ & $1.2$ & $84.9$ & $28.4$ & $54.7$ & $4.6$ & $61.5$ & $29.2$ & $59.6$ & $25.5$ & $54.7$ & $11.2$ & $36.3$
\\
SqSegV2~\cite{wu2019squeezesegv2} & $39.6$ & $82.7$ & $21.0$ & $22.6$ & $14.5$ & $15.9$ & $20.2$ & $24.3$ & $2.9$ & $88.5$ & $42.4$ & $65.5$ & $18.7$ & $73.8$ & $41.0$ & $68.5$ & $36.9$ & $58.9$ & $12.9$ & $41.0$
\\
RandLA-Net~\cite{hu2020randla} & $50.3$ & $94.0$ & $19.8$ & $21.4$ & $42.7$ & $38.7$ & $47.5$ & $48.8$ & $4.6$ & $90.4$ & $56.9$ & $67.9$ & $15.5$ & $81.1$ & $49.7$ & $78.3$ & $60.3$ & $59.0$ & $44.2$ & $38.1$
\\
RangeNet++~\cite{milioto2019rangenet++} & $52.2$ & $91.4$ & $25.7$ & $34.4$ & $25.7$ & $23.0$ & $38.3$ & $38.8$ & $4.8$ & $91.8$ & $65.0$ & $75.2$ & $27.8$ & $87.4$ & $58.6$ & $80.5$ & $55.1$ & $64.6$ & $47.9$ & $55.9$
\\
PolarNet~\cite{zhang2020polarnet} & $54.3$ & $93.8$ & $40.3$ & $30.1$ & $22.9$ & $28.5$ & $43.2$ & $40.2$ & $5.6$ & $90.8$ & $61.7$ & $74.4$ & $21.7$ & $90.0$ & $61.3$ & $84.0$ & $65.5$ & $67.8$ & $51.8$ & $57.5$
\\
MPF~\cite{alnaggar2021mpf} & $55.5$ & $93.4$ & $30.2$ & $38.3$ & $26.1$ & $28.5$ & $48.1$ & $46.1$ & $18.1$ & $90.6$ & $62.3$ & $74.5$ & $30.6$ & $88.5$ & $59.7$ & $83.5$ & $59.7$ & $69.2$ & $49.7$ & $58.1$
\\
3D-MiniNet~\cite{alonso20203dmininet} & $55.8$ & $90.5$ & $42.3$ & $42.1$ & $28.5$ & $29.4$ & $47.8$ & $44.1$ & $14.5$ & $91.6$ & $64.2$ & $74.5$ & $25.4$ & $89.4$ & $60.8$ & $82.8$ & $60.8$ & $66.7$ & $48.0$ & $56.6$
\\
SqSegV3~\cite{xu2020squeezesegv3} & $55.9$ & $92.5$ & $38.7$ & $36.5$ & $29.6$ & $33.0$ & $45.6$ & $46.2$ & $20.1$ & $91.7$ & $63.4$ & $74.8$ & $26.4$ & $89.0$ & $59.4$ & $82.0$ & $58.7$ & $65.4$ & $49.6$ & $58.9$
\\
KPConv~\cite{thomas2019kpconv} & $58.8$ & $96.0$ & $32.0$ & $42.5$ & $33.4$ & $44.3$ & $61.5$ & $61.6$ & $11.8$ & $88.8$ & $61.3$ & $72.7$ & $31.6$ & $95.0$ & $64.2$ & $84.8$ & $69.2$ & $69.1$ & $56.4$ & $47.4$
\\
SalsaNext~\cite{cortinhal2020salsanext} & $59.5$ & $91.9$ & $48.3$ & $38.6$ & $38.9$ & $31.9$ & $60.2$ & $59.0$ & $19.4$ & $91.7$ & $63.7$ & $75.8$ & $29.1$ & $90.2$ & $64.2$ & $81.8$ & $63.6$ & $66.5$ & $54.3$ & $62.1$
\\
FIDNet~\cite{zhao2021fidnet} & $59.5$ & $93.9$ & $54.7$ & $48.9$ & $27.6$ & $23.9$ & $62.3$ & $59.8$ & $23.7$ & $90.6$ & $59.1$ & $75.8$ & $26.7$ & $88.9$ & $60.5$ & $84.5$ & $64.4$ & $69.0$ & $53.3$ & $62.8$
\\
FusionNet~\cite{zhang2020fusionnet} & $61.3$ & $95.3$ & $47.5$ & $37.7$ & $41.8$ & $34.5$ & $59.5$ & $56.8$ & $11.9$ & $91.8$ & $68.8$ & $77.1$ & $30.8$ & $92.5$ & $69.4$ & $84.5$ & $69.8$ & $68.5$ & $60.4$ & $66.5$
\\
PCSCNet~\cite{park2023pcscnet} & $62.7$ & $95.7$ & $48.8$ & $46.2$ & $36.4$ & $40.6$ & $55.5$ & $68.4$ & $55.9$ & $89.1$ & $60.2$ & $72.4$ & $23.7$ & $89.3$ & $64.3$ & $84.2$ & $68.2$ & $68.1$ & $60.5$ & $63.9$
\\
KPRNet~\cite{kochanov2020kprnet} & $63.1$ & $95.5$ & $54.1$ & $47.9$ & $23.6$ & $42.6$ & $65.9$ & $65.0$ & $16.5$ & $93.2$ & \underline{$73.9$} & $80.6$ & $30.2$ & $91.7$ & $68.4$ & $85.7$ & $69.8$ & $71.2$ & $58.7$ & $64.1$
\\
TornadoNet~\cite{gerdzhev2021tornadonet} & $63.1$ & $94.2$ & $55.7$ & $48.1$ & $40.0$ & $38.2$ & $63.6$ & $60.1$ & $34.9$ & $89.7$ & $66.3$ & $74.5$ & $28.7$ & $91.3$ & $65.6$ & $85.6$ & $67.0$ & $71.5$ & $58.0$ & $65.9$
\\
LiteHDSeg~\cite{razani2021boundary} & $63.8$ & $92.3$ & $40.0$ & $55.4$ & $37.7$ & $39.6$ & $59.2$ & $71.6$ & $54.3$ & $93.0$ & $68.2$ & $78.3$ & $29.3$ & $91.5$ & $65.0$ & $78.2$ & $65.8$ & $65.1$ & $59.5$ & $67.7$
\\
RangeViT~\cite{ando2023rangevit} & $64.0$ & $95.4$ & $55.8$ & $43.5$ & $29.8$ & $42.1$ & $63.9$ & $58.2$ & $38.1$ & $93.1$ & $70.2$ & $80.0$ & $32.5$ & $92.0$ & $69.0$ & $85.3$ & $70.6$ & $71.2$ & $60.8$ & $64.7$
\\
CENet~\cite{cheng2022cenet} & $64.7$ & $91.9$ & $58.6$ & $50.3$ & $40.6$ & $42.3$ & $68.9$ & $65.9$ & $43.5$ & $90.3$ & $60.9$ & $75.1$ & $31.5$ & $91.0$ & $66.2$ & $84.5$ & $69.7$ & $70.0$ & $61.5$ & $67.6$
\\
SVASeg~\cite{zhao2022svaseg} & $65.2$ & $96.7$ & $56.4$ & $57.0$ & $49.1$ & $56.3$ & $70.6$ & $67.0$ & $15.4$ & $92.3$ & $65.9$ & $76.5$ & $23.6$ & $91.4$ & $66.1$ & $85.2$ & $72.9$ & $67.8$ & $63.9$ & $65.2$
\\
AMVNet~\cite{liong2020amvnet} & $65.3$ & $96.2$ & $59.9$ & $54.2$ & $48.8$ & $45.7$ & $71.0$ & $65.7$ & $11.0$ & $90.1$ & $71.0$ & $75.8$ & $32.4$ & $92.4$ & $69.1$ & $85.6$ & $71.7$ & $69.6$ & $62.7$ & $67.2$
\\
GFNet~\cite{qiu2022gfnet} & $65.4$ & $96.0$ & $53.2$ & $48.3$ & $31.7$ & $47.3$ & $62.8$ & $57.3$ & $44.7$ & $\mathbf{93.6}$ & $72.5$ & $\mathbf{80.8}$ & $31.2$ & $\mathbf{94.0}$ & $\mathbf{73.9}$ & $85.2$ & $71.1$ & $69.3$ & $61.8$ & $68.0$
\\
JS3C-Net~\cite{yan2021js3cnet} & $66.0$ & $95.8$ & $59.3$ & $52.9$ & $54.3$ & $46.0$ & $69.5$ & $65.4$ & $39.9$ & $88.9$ & $61.9$ & $72.1$ & $31.9$ & $92.5$ & $70.8$ & $84.5$ & $69.8$ & $67.9$ & $60.7$ & $68.7$
\\
MaskRange~\cite{gu2022maskrange} & $66.1$ & $94.2$ & $56.0$ & $55.7$ & $59.2$ & $52.4$ & $67.6$ & $64.8$ & $31.8$ & $91.7$ & $70.7$ & $77.1$ & $29.5$ & $90.6$ & $65.2$ & $84.6$ & $68.5$ & $69.2$ & $60.2$ & $66.6$ 
\\
SPVNAS~\cite{tang2020spvnas} & $66.4$ & $97.3$ & $51.5$ & $50.8$ & $59.8$ & $58.8$ & $65.7$ & $65.2$ & $43.7$ & $90.2$ & $67.6$ & $75.2$ & $16.9$ & $91.3$ & $65.9$ & $86.1$ & $73.4$ & $71.0$ & $64.2$ & $66.9$
\\
MSSNet~\cite{su2022mssnet} & $66.7$ & $96.8$ & $52.2$ & $48.5$ & $54.4$ & $56.3$ & $67.0$ & $70.9$ & $49.3$ & $90.1$ & $65.5$ & $74.9$ & $30.2$ & $90.5$ & $64.9$ & $84.9$ & $72.7$ & $69.2$ & $63.2$ & $65.1$
\\
Cylinder3D~\cite{zhu2021cylinder3d} & $68.9$ & $97.1$ & $67.6$ & $63.8$ & $50.8$ & $58.5$ & $73.7$ & $69.2$ & $48.0$ & $92.2$ & $65.0$ & $77.0$ & $32.3$ & $90.7$ & $66.5$ & $85.6$ & $72.5$ & $69.8$ & $62.4$ & $66.2$
\\
AF2S3Net~\cite{cheng2021af2s3net} & $69.7$ & $94.5$ & $65.4$ & $\mathbf{86.8}$ & $39.2$ & $41.1$ & $\mathbf{80.7}$ & $80.4$ & \underline{$74.3$} & $91.3$ & $68.8$ & $72.5$ & $\mathbf{53.5}$ & $87.9$ & $63.2$ & $70.2$ & $68.5$ & $53.7$ & $61.5$ & $71.0$
\\
RPVNet~\cite{xu2021rpvnet} & $70.3$  & \underline{$97.6$} & $68.4$ & $68.7$ & $44.2$ & $61.1$ & $75.9$ & $74.4$ & $73.4$ & \underline{$93.4$} & $70.3$ & \underline{$80.7$} & $33.3$ & $93.5$ & $72.1$ & $86.5$ & \underline{$75.1$} & $71.7$ & $64.8$ & $61.4$
\\
SDSeg3D~\cite{li2022sdseg3d} & $70.4$ & $97.4$ & $58.7$ & $54.2$ & $54.9$ & $65.2$ & $70.2$ & $74.4$ & $52.2$ & $90.9$ & $69.4$ & $76.7$ & $41.9$ & $93.2$ & $71.1$ & $86.1$ & $74.3$ & $71.1$ & $65.4$ & $70.6$
\\
GASN~\cite{ye2022gasn} & $70.7$ & $96.9$ & $65.8$ & $58.0$ & $59.3$ & $61.0$ & \underline{$80.4$} & $\mathbf{82.7}$ & $46.3$ & $89.8$ & $66.2$ & $74.6$ & $30.1$ & $92.3$ & $69.6$ & \underline{$87.3$} & $73.0$ & $72.5$ & $66.1$ & \underline{$71.6$}
\\
PVKD~\cite{hou2022pvkd} & $71.2$ & $97.0$ & $67.9$ & $69.3$ & $53.5$ & $60.2$ & $75.1$ & $73.5$ & $50.5$ & $91.8$ & $70.9$ & $77.5$ & $41.0$ & $92.4$ & $69.4$ & $86.5$ & $73.8$ & $71.9$ & $64.9$ & $65.8$
\\
\rowcolor{fr_purple!8}\textbf{Fast-FRNet} & $72.5$ & $97.1$ & $66.0$ & $73.1$ & $59.3$ & $65.7$ & $76.0$ & $78.9$ & $54.5$ & $91.8$ & $72.6$ & $77.6$ & $42.0$ & $92.4$ & $70.6$ & $86.5$ & $71.9$ & $72.5$ & $63.1$ & $66.7$
\\
2DPASS~\cite{yan20222dpass} & $72.9$ & $97.0$ & $63.6$ & $63.4$ & \underline{$61.1$} & $61.5$ & $77.9$ & \underline{$81.3$} & $74.1$ & $89.7$ & $67.4$ & $74.7$ & $40.0$ & $93.5$ & \underline{$72.9$} & $86.2$ & $73.9$ & $71.0$ & $65.0$ & $70.4$
\\
RangeFormer~\cite{kong2023rangeformer} & $73.3$ & $96.7$ & $69.4$ & $73.7$ & $59.9$ & $66.2$ & $78.1$ & $75.9$ & $58.1$ & $92.4$ & $73.0$ & $78.8$ & $42.4$ & $92.3$ & $70.1$ & $86.6$ & $73.3$ & \underline{$72.8$} & $66.4$ & $66.6$
\\
\rowcolor{fr_purple!8}\textbf{FRNet} & $73.3$ & $97.3$ & $67.9$ & $74.6$ & $59.4$ & $66.3$ & $78.1$ & $79.2$ & $57.3$ & $92.1$ & $73.0$ & $78.1$ & $41.8$ & $92.7$ & $71.0$ & $86.7$ & $73.2$ & $72.5$ & $64.7$ & $67.3$
\\
SphereFormer~\cite{lai2023sphereformer} & \underline{$74.8$} & $97.5$ & \underline{$70.1$} & $70.5$ & $59.6$ & \underline{$67.7$} & $79.0$ & $80.4$ & $\mathbf{75.3}$ & $91.8$ & $69.7$ & $78.2$ & $41.3$ & \underline{$93.8$} & $72.8$ & $86.7$ & \underline{$75.1$} & $72.4$ & \underline{$66.8$} & $\mathbf{72.9}$
\\
UniSeg~\cite{liu2023uniseg} & $\mathbf{75.2}$ & $\mathbf{97.9}$ & $\mathbf{71.9}$ & \underline{$75.2$} & $\mathbf{63.6}$ & $\mathbf{74.1}$ & $78.9$ & $74.8$ & $60.6$ & $92.6$ & $\mathbf{74.0}$ & $79.5$ & \underline{$46.1$} & $93.4$ & $72.7$ & $\mathbf{87.5}$ & $\mathbf{76.3}$ & $\mathbf{73.1}$ & $\mathbf{68.3}$ & $68.5$
\\\bottomrule
\end{tabular}
\end{adjustbox}
\label{table:semantickitti-class}
\end{table*}

\begin{table*}[t]
\caption{\textbf{The class-wise IoU scores} of different LiDAR semantic segmentation approaches on the \textit{val} set of nuScenes~\cite{fong2022panoptic}. All IoU scores are given in percentage (\%). The \textbf{best} and \underline{second best} scores for each class are highlighted in \textbf{bold} and \underline{underline}.}
\vspace{-0.1cm}
\begin{adjustbox}{width=\textwidth}
\begin{tabular}{r|c|cccccccccccccccc}
\toprule
\textbf{Method~\small{(year)}} & \rotatebox{90}{mIoU} & \rotatebox{90}{barrier} & \rotatebox{90}{bicycle} & \rotatebox{90}{bus} & \rotatebox{90}{car} & \rotatebox{90}{construction-vehicle~} & \rotatebox{90}{motorcycle} & \rotatebox{90}{pedestrian} & \rotatebox{90}{traffic-cone} & \rotatebox{90}{trailer} & \rotatebox{90}{truck} & \rotatebox{90}{driveable-surface} & \rotatebox{90}{other-ground} & \rotatebox{90}{sidewalk} & \rotatebox{90}{terrain} & \rotatebox{90}{manmade} & \rotatebox{90}{vegetation}
\\\midrule
AF2S3Net~\cite{cheng2021af2s3net}~\small{['21]} & $62.2$ & $60.3$ & $12.6$ & $82.3$ & $80.0$ & $20.1$ & $62.0$ & $59.0$ & $49.0$ & $42.2$ & $67.4$ & $94.2$ & $68.0$ & $64.1$ & $68.6$ & $82.9$ & $82.4$
\\
RangeNet++~\cite{milioto2019rangenet++}~\small{['19]} & $65.5$ & $66.0$ & $21.3$ & $77.2$ & $80.9$ & $30.2$ & $66.8$ & $69.6$ & $52.1$ & $54.2$ & $72.3$ & $94.1$ & $66.6$ & $63.5$ & $70.1$ & $83.1$ & $79.8$
\\
PolarNet~\cite{zhang2020polarnet}~\small{['20]} & $71.0$ & $74.7$ & $28.2$ & $85.3$ & $90.9$ & $35.1$ & $77.5$ & $71.3$ & $58.8$ & $57.4$ & $76.1$ & $96.5$ & $71.1$ & $74.7$ & $74.0$ & $87.3$ & $85.7$
\\
PCSCNet~\cite{park2023pcscnet}~\small{['22]} & $72.0$ & $73.3$ & $42.2$ & $87.8$ & $86.1$ & $44.9$ & $82.2$ & $76.1$ & $62.9$ & $49.3$ & $77.3$ & $95.2$ & $66.9$ & $69.5$ & $72.3$ & $83.7$ & $82.5$
\\
SalsaNext~\cite{cortinhal2020salsanext}~\small{['20]} & $72.2$ & $74.8$ & $34.1$ & $85.9$ & $88.4$ & $42.2$ & $72.4$ & $72.2$ & $63.1$ & $61.3$ & $76.5$ & $96.0$ & $70.8$ & $71.2$ & $71.5$ & $86.7$ & $84.4$
\\
SVASeg~\cite{zhao2022svaseg}~\small{['22]} & $74.7$ & $73.1$ & \underline{$44.5$} & $88.4$ & $86.6$ & $48.2$ & $80.5$ & $77.7$ & $65.6$ & $57.5$ & $82.1$ & $96.5$ & $70.5$ & $74.7$ & $74.6$ & $87.3$ & $86.9$
\\
RangeViT~\cite{ando2023rangevit}~\small{['23]} & $75.2$ & $75.5$ & $40.7$ & $88.3$ & $90.1$ & $49.3$ & $79.3$ & $77.2$ & $66.3$ & $65.2$ & $80.0$ & $96.4$ & $71.4$ & $73.8$ & $73.8$ & $89.9$ & $87.2$
\\
Cylinder3D~\cite{zhu2021cylinder3d}~\small{['21]} & $76.1$ & $76.4$ & $40.3$ & $91.2$ & $\mathbf{93.8}$ & $51.3$ & $78.0$ & $78.9$ & $64.9$ & $62.1$ & $84.4$ & $96.8$ & $71.6$ & $76.4$ & $75.4$ & $90.5$ & $87.4$
\\
AMVNet~\cite{liong2020amvnet}~\small{['20]} & $76.1$ & $\mathbf{79.8}$ & $32.4$ & $82.2$ & $86.4$ & $\mathbf{62.5}$ & $81.9$ & $75.3$ & $\mathbf{72.3}$ & $\mathbf{83.5}$ & $65.1$ & $\mathbf{97.4}$ & $67.0$ & $\mathbf{78.8}$ & $74.6$ & \underline{$90.8$} & $87.9$
\\
RPVNet~\cite{xu2021rpvnet}~\small{['21]} & $77.6$ & $78.2$ & $43.4$ & $92.7$ & \underline{$93.2$} & $49.0$ & $85.7$ & \underline{$80.5$} & $66.0$ & $66.9$ & $84.0$ & $96.9$ & $73.5$ & $75.9$ & $70.6$ & $90.6$ & \underline{$88.9$}
\\
RangeFormer~\cite{kong2023rangeformer}~\small{['23]} & $78.1$ & $78.0$ & $\mathbf{45.2}$ & $94.0$ & $92.9$ & $58.7$ & $83.9$ & $77.9$ & \underline{$69.1$} & $63.7$ & \underline{$85.6$} & $96.7$ & $74.5$ & $75.1$ & $75.3$ & $89.1$ & $87.5$
\\
SphereFormer~\cite{lai2023sphereformer}~\small{['23]} & $78.4$ & $77.7$ & $43.8$ & $94.5$ & $93.1$ & $52.4$ & $\mathbf{86.9}$ & $\mathbf{81.2}$ & $65.4$ & \underline{$73.4$} & $85.3$ & $97.0$ & $73.4$ & $75.4$ & $75.0$ & $\mathbf{91.0}$ & $\mathbf{89.2}$
\\
\rowcolor{fr_purple!8}\textbf{Fast-FRNet} & \underline{$78.8$} & \underline{$78.7$} & $42.3$ & $\mathbf{95.6}$ & $93.1$ & \underline{$58.9$} & \underline{$86.3$} & $77.9$ & $66.9$ & $72.1$ & $85.4$ & $97.0$ & $\mathbf{76.3}$ & $76.5$ & \underline{$76.2$} & $89.7$ & $87.8$
\\
\rowcolor{fr_purple!8}\textbf{FRNet} & $\mathbf{79.0}$ & $78.5$ & $43.9$ & \underline{$95.4$} & \underline{$93.2$} & $56.3$ & $85.8$ & $79.0$ & $68.5$ & $72.8$ & $\mathbf{86.5}$ & \underline{$97.1$} & \underline{$75.9$} & \underline{$77.0$} & $\mathbf{76.4}$ & $89.7$ & $88.0$
\\\bottomrule
\end{tabular}
\end{adjustbox}
\label{table:nuscenes-class-val}
\vspace{0.1cm}
\end{table*}

\begin{table*}[t]
\caption{\textbf{The class-wise IoU scores} of different LiDAR semantic segmentation approaches on the \textit{test} set of nuScenes~\cite{fong2022panoptic}. All IoU scores are given in percentage (\%). The \textbf{best} and \underline{second best} scores for each class are highlighted in \textbf{bold} and \underline{underline}.}
\vspace{-0.1cm}
\begin{adjustbox}{width=\textwidth}
\begin{tabular}{r|c|cccccccccccccccc}
\toprule
\textbf{Method~\small{(year)}} & \rotatebox{90}{mIoU} & \rotatebox{90}{barrier} & \rotatebox{90}{bicycle} & \rotatebox{90}{bus} & \rotatebox{90}{car} & \rotatebox{90}{construction-vehicle~} & \rotatebox{90}{motorcycle} & \rotatebox{90}{pedestrian} & \rotatebox{90}{traffic-cone} & \rotatebox{90}{trailer} & \rotatebox{90}{truck} & \rotatebox{90}{driveable-surface} & \rotatebox{90}{other-ground} & \rotatebox{90}{sidewalk} & \rotatebox{90}{terrain} & \rotatebox{90}{manmade} & \rotatebox{90}{vegetation}
\\\midrule
PolarNet~\cite{zhang2020polarnet}~\small{['20]} & $69.4$ & $72.2$ & $16.8$ & $77.0$ & $86.5$ & $51.1$ & $69.7$ & $64.8$ & $54.1$ & $69.7$ & $63.5$ & $96.6$ & $67.1$ & $77.7$ & $72.1$ & $87.1$ & $84.5$
\\
JS3C-Net~\cite{yan2021js3cnet}~\small{['21]} & $73.6$ & $80.1$ & $26.2$ & $87.8$ & $84.5$ & $55.2$ & $72.6$ & $71.3$ & $66.3$ & $76.8$ & $71.2$ & $96.8$ & $64.5$ & $76.9$ & $74.1$ & $87.5$ & $86.1$
\\
PMF~\cite{zhuang2021pmf}~\small{['21]} & $77.0$ & $82.0$ & $40.0$ & $81.0$ & $88.0$ & $64.0$ & $79.0$ & $80.0$ & $76.0$ & $81.0$ & $67.0$ & $97.0$ & $68.0$ & $78.0$ & $74.0$ & $90.0$ & $88.0$
\\
Cylinder3D~\cite{zhu2021cylinder3d}~\small{['21]} & $77.2$ & $82.8$ & $29.8$ & $84.3$ & $89.4$ & $63.0$ & $79.3$ & $77.2$ & $73.4$ & $84.6$ & $69.1$ & $97.7$ & $70.2$ & $80.3$ & $75.5$ & $90.4$ & $87.6$
\\
AMVNet~\cite{liong2020amvnet}~\small{['20]} & $77.3$ & $80.6$ & $32.0$ & $81.7$ & $88.9$ & $67.1$ & $84.3$ & $76.1$ & $73.5$ & $84.9$ & $67.3$ & $97.5$ & $67.4$ & $79.4$ & $75.5$ & $91.5$ & $88.7$
\\
SPVCNN~\cite{tang2020spvnas}~\small{['20]} & $77.4$ & $80.0$ & $30.0$ & $91.9$ & $90.8$ & $64.7$ & $79.0$ & $75.6$ & $70.9$ & $81.0$ & $74.6$ & $97.4$ & $69.2$ & $80.0$ & $76.1$ & $89.3$ & $87.1$
\\
AF2S3Net~\cite{cheng2021af2s3net}~\small{['21]} & $78.3$ & $78.9$ & $52.2$ & $89.9$ & $84.2$ & $77.4$ & $74.3$ & $77.3$ & $72.0$ & $83.9$ & $73.8$ & $97.1$ & $66.5$ & $77.5$ & $74.0$ & $87.7$ & $86.8$
\\
2D3DNet~\cite{genova20212d3dnet}~\small{['21]} & $80.0$ & $83.0$ & $59.4$ & $88.0$ & $85.1$ & $63.7$ & $84.4$ & $82.0$ & $76.0$ & $84.8$ & $71.9$ & $96.9$ & $67.4$ & $79.8$ & $76.0$ & \underline{$92.1$} & $89.2$
\\
RangeFormer~\cite{kong2023rangeformer}~\small{['23]} & $80.1$ & $85.6$ & $47.4$ & $91.2$ & $90.9$ & $70.7$ & $84.7$ & $77.1$ & $74.1$ & $83.2$ & $72.6$ & $97.5$ & $70.7$ & $79.2$ & $75.4$ & $91.3$ & $88.9$
\\
GASN~\cite{ye2022gasn}~\small{['22]} & $80.4 $ & $85.5$ & $43.2$ & $90.5$ & \underline{$92.1$} & $64.7$ & $86.0$ & $83.0$ & $73.3$ & $83.9$ & $75.8$ & $97.0$ & $71.0$ & \underline{$81.0$} & $\mathbf{77.7}$ & $91.6$ & $\mathbf{90.2}$
\\
2DPASS~\cite{yan20222dpass}~\small{['22]} & $80.8$ & $81.7$ & $55.3$ & $92.0$ & $91.8$ & $73.3$ & $86.5$ & $78.5$ & $72.5$ & $84.7$ & $75.5$ & $97.6$ & $69.1$ & $79.9$ & $75.5$ & $90.2$ & $88.0$
\\
LidarMultiNet~\cite{ye2023lidarmultinet}~\small{['23]} & $81.4$ & $80.4$ & $48.4$ & \underline{$94.3$} & $90.0$ & $71.5$ & $87.2$ & $\mathbf{85.2}$ & $\mathbf{80.4}$ & \underline{$86.9$} & $74.8$ & \underline{$97.8$} & $67.3$ & $80.7$ & $76.5$ & \underline{$92.1$} & \underline{$89.6$}
\\
SphereFormer~\cite{lai2023sphereformer}~\small{['23]} & $81.9$ & $83.3$ & $39.2$ & $\mathbf{94.7}$ & $\mathbf{92.5}$ & \underline{$77.5$} & $84.2$ & \underline{$84.4$} & \underline{$79.1$} & $\mathbf{88.4}$ & $\mathbf{78.3}$ & $\mathbf{97.9}$ & $69.0$ & $\mathbf{81.5}$ & \underline{$77.2$} & $\mathbf{93.4}$ & $\mathbf{90.2}$
\\
\rowcolor{fr_purple!8}\textbf{Fast-FRNet} & $82.1$ & $85.3$ & $62.9$ & $92.1$ & $91.5$ & $76.5$ & $87.7$ & $75.9$ & $73.2$ & $86.2$ & $76.0$ & \underline{$97.8$} & $\mathbf{71.9}$ & $80.9$ & $77.1$ & $90.8$ & $88.1$
\\
\rowcolor{fr_purple!8}\textbf{FRNet} & \underline{$82.5$} & \underline{$85.8$} & \underline{$65.4$} & $92.1$ & $91.6$ & $77.4$ & \underline{$87.9$} & $77.4$ & $74.3$ & $86.0$ & $75.7$ & \underline{$97.8$} & \underline{$71.8$} & $80.8$ & $77.0$ & $91.0$ & $88.3$
\\
UniSeg~\cite{liu2023uniseg}~\small{['23]} & $\mathbf{83.5}$ & $\mathbf{85.9}$ & $\mathbf{71.2}$ & $92.1$ & $91.6$ & $\mathbf{80.5}$ & $\mathbf{88.0}$ & $80.9$ & $76.0$ & $86.3$ & \underline{$76.7$} & $97.7$ & \underline{$71.8$} & $80.7$ & $76.7$ & $91.3$ & $88.8$
\\\bottomrule
\end{tabular}
\end{adjustbox}
\label{table:nuscenes-class-test}
\end{table*}

\begin{table*}[t]
\caption{\textbf{The class-wise IoU scores} of different LiDAR semantic segmentation approaches on the ScribbleKITTI~\cite{unal2022scribblekitti} leaderboard. All IoU scores are given in percentage (\%). The \textbf{best} and \underline{second best} scores for each class are highlighted in \textbf{bold} and \underline{underline}.}
\vspace{-0.1cm}
\begin{adjustbox}{width=\textwidth}
\begin{tabular}{r|c|ccccccccccccccccccc}
\toprule
\textbf{Method} & \rotatebox{90}{$\text{mIoU}$} & \rotatebox{90}{car} & \rotatebox{90}{bicycle} & \rotatebox{90}{motorcycle} & \rotatebox{90}{truck} & \rotatebox{90}{other-vehicle~} & \rotatebox{90}{person} & \rotatebox{90}{bicyclist} & \rotatebox{90}{motorcyclist} & \rotatebox{90}{road} & \rotatebox{90}{parking} & \rotatebox{90}{sidewalk} & \rotatebox{90}{other-ground~} & \rotatebox{90}{building} & \rotatebox{90}{fence} & \rotatebox{90}{vegetation} & \rotatebox{90}{trunk} & \rotatebox{90}{terrain} & \rotatebox{90}{pole} & \rotatebox{90}{traffic-sign}
\\\midrule\midrule
RangeNet++~\cite{milioto2019rangenet++} & $44.6$ & $84.6$ & $24.6$ & $38.9$ & $13.7$ & $12.9$ & $29.0$ & $51.4$ & $0.0$ & $86.0$ & \underline{$35.2$} & $72.4$ & \underline{$4.5$} & $80.2$ & $41.6$ & $80.1$ & $51.2$ & $66.9$ & $43.9$ & $34.5$
\\
SalsaNext~\cite{cortinhal2020salsanext} & $50.3$ & $85.6$ & $33.8$ & $42.7$ & $8.4$ & $20.8$ & $64.3$ & $71.5$ & \underline{$0.2$} & $85.6$ & $31.6$ & $72.0$ & $1.1$ & $84.0$ & $39.9$ & $81.0$ & $62.0$ & $63.6$ & $60.5$ & $46.3$
\\
RangeViT~\cite{ando2023rangevit} & $53.6$ & $85.6$ & $31.6$ & $50.1$ & $40.3$ & $36.3$ & $57.6$ & $68.7$ & $0.0$ & $86.1$ & $32.6$ & $75.2$ & $0.3$ & $87.9$ & $49.3$ & $83.6$ & $62.8$ & $67.5$ & $59.6$ & $43.7$
\\
FIDNet~\cite{zhao2021fidnet} & $54.1$ & $85.6$ & $36.7$ & $48.7$ & $60.8$ & $38.4$ & $63.3$ & $68.2$ & $0.0$ & $84.1$ & $25.9$ & $71.2$ & $0.4$ & $85.6$ & $41.3$ & $81.7$ & $64.1$ & $62.7$ & $61.5$ & $48.0$
\\
MinkNet~\cite{choy2019minkunet} & $55.0$ & $88.1$ & $13.2$ & $55.1$ & $72.3$ & $36.9$ & $61.3$ & $77.1$ & $0.0$ & $83.4$ & $32.7$ & $71.0$ & $0.3$ & $90.0$ & $50.0$ & $84.1$ & \underline{$66.6$} & $65.8$ & $61.6$ & $35.2$
\\
CENet~\cite{cheng2022cenet} & $55.7$ & $86.1$ & $39.4$ & $53.2$ & $61.0$ & $46.1$ & $69.2$ & $72.2$ & $0.0$ & $85.7$ & $28.7$ & $72.6$ & $1.1$ & $85.8$ & $43.1$ & $81.8$ & $64.2$ & $63.8$ & $59.6$ & $45.0$
\\
SPVCNN~\cite{tang2020spvnas} & $56.9$ & $88.6$ & $25.7$ & $55.9$ & $67.4$ & $48.8$ & $65.0$ & $78.2$ & $0.0$ & $82.6$ & $30.4$ & $70.1$ & $0.3$ & $\mathbf{90.5}$ & $49.6$ & $84.4$ & $\mathbf{67.6}$ & $66.1$ & $61.6$ & $\mathbf{48.7}$
\\
Cylinder3D~\cite{zhu2021cylinder3d} & $57.0$ & $88.5$ & $39.9$ & $58.0$ & $58.4$ & $48.1$ & $68.6$ & $77.0$ & $\mathbf{0.5}$ & $84.4$ & $30.4$ & $72.2$ & $2.5$ & $89.4$ & $48.4$ & $81.9$ & $64.6$ & $59.8$ & $61.2$ & $\mathbf{48.7}$
\\
\rowcolor{fr_purple!8}\textbf{Fast-FRNet} & $62.4$ & \underline{$90.8$} & $41.0$ & $\mathbf{66.8}$ & \underline{$81.7$} & $\mathbf{64.0}$ & $70.5$ & \underline{$84.2$} & $0.0$ & $91.3$ & $35.1$ & $78.3$ & $0.0$ & $89.4$ & \underline{$64.4$} & $85.0$ & $65.5$ & $69.5$ & $60.5$ & $47.7$
\\
RangeFormer~\cite{kong2023rangeformer} & \underline{$63.0$} & $\mathbf{92.6}$ & $\mathbf{51.6}$ & \underline{$65.7$} & $74.4$ & $49.6$ & \underline{$71.6$} & $82.1$ & $0.0$ & $\mathbf{94.8}$ & $\mathbf{44.4}$ & $\mathbf{80.6}$ & $\mathbf{11.4}$ & $85.6$ & $56.9$ & $\mathbf{87.2}$ & $64.1$ & $\mathbf{77.0}$ & $\mathbf{62.7}$ & $45.1$
\\
\rowcolor{fr_purple!8}\textbf{FRNet} & $\mathbf{63.1}$ & $90.5$ & \underline{$42.3$} & $\mathbf{66.8}$ & $\mathbf{82.4}$ & \underline{$63.0$} & $\mathbf{73.4}$ & $\mathbf{86.2}$ & $0.0$ & \underline{$92.1$} & $35.1$ & \underline{$79.1$} & $0.3$ & \underline{$90.4$} & $\mathbf{64.7}$ & \underline{$85.4$} & $65.8$ & \underline{$70.7$} & \underline{$61.8$} & \underline{$48.6$}
\\\bottomrule
\end{tabular}
\end{adjustbox}
\label{table:scribblekitti-class}
\vspace{0.1cm}
\end{table*}

\begin{table*}[t]
\caption{\textbf{The class-wise IoU scores} of different LiDAR semantic segmentation approaches on the SemanticPOSS~\cite{pan2020semanticposs} leaderboard. All IoU scores are given in percentage (\%). The \textbf{best} and \underline{second best} scores for each class are highlighted in \textbf{bold} and \underline{underline}.}
\vspace{-0.1cm}
\centering\scalebox{1.0}{
\begin{tabular}{r|c|ccccccccccccc}
\toprule
\textbf{Method~\small{(year)}} & \rotatebox{90}{$\text{mIoU}$} & \rotatebox{90}{person} & \rotatebox{90}{rider} & \rotatebox{90}{car} & \rotatebox{90}{truck} & \rotatebox{90}{plants} & \rotatebox{90}{traffic sign~} & \rotatebox{90}{pole} & \rotatebox{90}{trashcan} & \rotatebox{90}{building} & \rotatebox{90}{cone/stone~} & \rotatebox{90}{walk} & \rotatebox{90}{fence} & \rotatebox{90}{bike}
\\\midrule\midrule
SqSeg~\cite{wu2018squeezeseg}~\small{['18]} & $18.9$ & $14.2$ & $1.0$ & $13.2$ & $10.4$ & $28.0$ & $5.1$ & $5.7$ & $2.3$ & $43.6$ & $0.2$ & $15.6$ & $31.0$ & $75.0$
\\
SqSegV2~\cite{wu2019squeezesegv2}~\small{['19]} & $30.0$ & $48.0$ & $9.4$ & $48.5$ & $11.3$ & $50.1$ & $6.7$ & $6.2$ & $14.8$ & $60.4$ & $5.2$ & $22.1$ & $36.1$ & $71.3$
\\
RangeNet++~\cite{milioto2019rangenet++}~\small{['19]} & $30.9$ & $57.3$ & $4.6$ & $35.0$ & $14.1$ & $58.3$ & $3.9$ & $6.9$ & $24.1$ & $66.1$ & $6.6$ & $23.4$ & $28.6$ & $73.5$
\\
MINet~\cite{li2021minet}~\small{['21]} & $43.2$ & $62.4$ & $12.1$ & $63.8$ & $22.3$ & $68.6$ & $16.7$ & $30.1$ & $28.9$ & $75.1$ & $28.6$ & $32.2$ & $44.9$ & $76.3$
\\
FIDNet~\cite{zhao2021fidnet}~\small{['21]} & $46.4$ & $72.2$ & $23.1$ & $72.7$ & $23.0$ & $68.0$ & $22.2$ & $28.6$ & $16.3$ & $73.1$ & $34.0$ & $40.9$ & $50.3$ & $79.1$
\\
SPVCNN~\cite{tang2020spvnas}~\small{['20]} & $48.4$ & $72.5$ & $24.7$ & $72.1$ & \underline{$31.4$} & $72.7$ & $10.8$ & \underline{$41.3$} & $31.8$ & $78.4$ & $23.8$ & $42.6$ & $51.7$ & $75.3$
\\
CENet~\cite{cheng2022cenet}~\small{['22]} & $50.3$ & $75.5$ & $22.0$ & $77.6$ & $25.3$ & $72.2$ & $18.2$ & $31.5$ & $\mathbf{48.1}$ & $76.3$ & $27.7$ & $47.7$ & $51.4$ & $\mathbf{80.3}$
\\
\rowcolor{fr_purple!8}\textbf{Fast-FRNet} & $52.5$ & $76.9$ & $28.3$ & $79.9$ & $28.8$ & $73.8$ & \underline{$30.2$} & $32.9$ & $28.5$ & $80.9$ & \underline{$40.7$} & $47.3$ & $55.9$ & $79.0$
\\
Cylinder3D~\cite{zhu2021cylinder3d}~\small{['21]} & $52.9$ & $75.9$ & $\mathbf{30.0}$ & $75.8$ & $28.7$ & $\mathbf{75.7}$ & $29.5$ & $37.2$ & \underline{$36.7$} & $\mathbf{82.3}$ & $34.1$ & $47.5$ & $53.9$ & \underline{$80.1$}
\\
MinkNet~\cite{choy2019minkunet}~\small{['19]} & \underline{$53.1$} & $\mathbf{77.8}$ & \underline{$29.1$} & $\mathbf{81.3}$ & $\mathbf{33.9}$ & \underline{$75.2$} & $22.0$ & $\mathbf{42.5}$ & $36.4$ & $80.7$ & $23.9$ & $\mathbf{51.2}$ & $\mathbf{57.5}$ & $79.1$
\\
\rowcolor{fr_purple!8}\textbf{FRNet} & $\mathbf{53.5}$ & $\underline{77.7}$ & $28.7$ & \underline{$81.2$} & $28.9$ & $74.3$ & $\mathbf{30.4}$ & $34.5$ & $29.8$ & \underline{$81.1$} & $\mathbf{45.6}$ & \underline{$47.9$} & \underline{$56.4$} & $79.2$
\\\bottomrule
\end{tabular}}
\label{table:semanticposs-class}
\end{table*}

\subsection{Per-Class Results}

In this section, we supplement the complete (class-wise) segmentation results of the baselines, prior works, and our proposed FRNet and Fast-FRNet.

\subsubsection{SemanticKITTI}

\cref{table:semantickitti-class} shows the class-wise IoU scores among different LiDAR semantic segmentation methods on the \textit{test} set of SemanticKITTI~\cite{behley2019semantickitti}. We compare FRNet with different LiDAR representations, including point-view, range-view, sparse-voxel-view, and multi-view methods. Results demonstrate that FRNet achieves appealing performance among state-of-the-art methods. Compared with SphereFormer~\cite{lai2023sphereformer} and UniSeg~\cite{liu2023uniseg}, although FRNet obtains sub-par performance, it maintains satisfactory efficiency for real-time processing, achieving the balance between efficiency and accuracy. More results about efficiency and accuracy can be found in the main body. It is worth noting that Fast-FRNet achieves the fastest speed of inference with only a little sacrifice of performance.

\subsubsection{nuScenes}

\cref{table:nuscenes-class-val} and \cref{table:nuscenes-class-test} show the class-wise IoU scores among different LiDAR semantic segmentation methods on the $val$ and $test$ set of nuScenes~\cite{fong2022panoptic}, respectively. The results demonstrate the great advantages of Fast-FRNet and FRNet. For the much sparser LiDAR points, the performance of Fast-FRNet is very close to FRNet with fewer parameters. Compared with popular range-view methods, FRNet achieves great improvement in some dynamic instances, such as \textit{car}, \textit{bicycle}, \textit{motorcycle}, \textit{etc}.

\subsubsection{ScribbleKITTI}

To prove the great advantages of FRNet among range-view methods, we also reimplement the popular LiDAR segmentation methods on ScribbleKITTI~\cite{unal2022scribblekitti}, including RangeNet++~\cite{milioto2019rangenet++}, SalsaNext~\cite{cortinhal2020salsanext}, FIRNet~\cite{zhao2021fidnet}, CENet~\cite{cheng2022cenet}, and RangeViT~\cite{ando2023rangevit}. All the models are trained with their official settings. As shown in \cref{table:scribblekitti-class}, FRNet achieves state-of-the-art performance. However, the weakly-annotated labels make the frustum-level pseudo labels largely covered by unannotated labels, which will limit performance improvement, especially for \textit{other-ground}.

\subsubsection{SemanticPOSS}

We also reimplement some popular sparse-voxel-view methods, including SPVCNN~\cite{tang2020spvnas}, Cylinder3D~\cite{zhu2021cylinder3d}, and MinkNet~\cite{choy2019minkunet}, on SemanticPOSS~\cite{pan2020semanticposs}. All the experiments are conducted based on MMDetection3D framework~\cite{mmdet3d}. \cref{table:semanticposs-class} summarizes the class-wise IoU scores among these methods and some other popular range-view methods. Results demonstrate that FRNet achieves state-of-the-art performance among both range-view and voxel-view methods. For some tiny objects, such as \textit{traffic sign} and \textit{cone/stone}, FRNet achieves superior improvements.

\section{Broader Impact}

In this section, we elaborate on the positive societal influence and potential limitations of the proposed FRNet.

\subsection{Positive Societal Impacts}

Conducting efficient and accurate range view LiDAR semantic segmentation for autonomous driving has several positive societal impacts, particularly in terms of enhancing road safety, improving transportation efficiency, and contributing to broader technological advancements.

\begin{itemize}
    \item \textbf{Enhanced Road Safety.} One of the most significant benefits is the improvement in road safety. Autonomous vehicles equipped with advanced LiDAR semantic segmentation can accurately detect and classify objects in their environment, such as other vehicles, pedestrians, cyclists, and road obstacles. This precise perception capability allows for safer navigation and decision-making, potentially reducing accidents caused by human error.

    \item \textbf{Reduced Traffic Congestion.} Autonomous vehicles with advanced perception systems can optimize driving patterns, leading to smoother traffic flow. This can significantly reduce traffic congestion, especially in urban areas, thereby saving time for commuters and reducing stress associated with driving.

    \item \textbf{Emergency Response and Healthcare.} In emergency situations, autonomous vehicles can be used to quickly and safely transport patients or deliver medical supplies. The precision of LiDAR segmentation ensures that these vehicles can navigate through complex environments effectively.

    \item \textbf{Advancements in Smart City Infrastructure.} The integration of autonomous vehicles with smart city initiatives can lead to more efficient urban planning and infrastructure development. LiDAR data can be used not just for navigation but also for gathering urban data, which can inform city planning and maintenance.
\end{itemize}

\subsection{Potential Limitations}

Although the proposed FRNet and Fast-FRNet are capable of providing a better trade-off between LiDAR segmentation network efficiency and accuracy, there are several limitations within the current framework. First, frustum-level supervision counts the high-frequency semantic labels in the frustum region, which will cover the objects with few points. Such supervision weakens the performance of FRNet on tiny objects. Second, FRNet cannot handle objects with similar structures well. Some objects share a similar appearance with different semantic attributes, which limits the ability of FRNet to distinguish them in a 2D manner.

\section{Public Resources Used}
\label{sec:public-resources-used}

In this section, we acknowledge the use of public resources, during the course of this work.

\subsection{Public Codebase Used}

We acknowledge the use of the following public codebase, during the course of this work:

\begin{itemize}
    \item MMDetection3D\footnote{\url{https://github.com/open-mmlab/mmdetection3d}.} \dotfill Apache License 2.0
    \item MMEngine\footnote{\url{https://github.com/open-mmlab/mmengine}.} \dotfill Apache License 2.0
    \item MMCV\footnote{\url{https://github.com/open-mmlab/mmcv}.} \dotfill Apache License 2.0
    \item MMDetection\footnote{\url{https://github.com/open-mmlab/mmdetection}.} \dotfill Apache License 2.0
    \item PCSeg\footnote{\url{https://github.com/PJLab-ADG/PCSeg}.} \dotfill Apache License 2.0
    \item Pointcept\footnote{\url{https://github.com/Pointcept/Pointcept}.}\dotfill MIT License
\end{itemize}

\subsection{Public Datasets Used}

We acknowledge the use of the following public datasets, during the course of this work:

\begin{itemize}
    \item SemanticKITTI\footnote{\url{http://semantic-kitti.org}.} \dotfill CC BY-NC-SA 4.0
    \item SemanticKITTI-API\footnote{\url{https://github.com/PRBonn/semantic-kitti-api}.} \dotfill MIT License
    \item nuScenes\footnote{\url{https://www.nuscenes.org/nuscenes}.} \dotfill CC BY-NC-SA 4.0
    \item nuScenes-devkit\footnote{\url{https://github.com/nutonomy/nuscenes-devkit}.} \dotfill Apache License 2.0
    \item ScribbleKITTI\footnote{\url{https://github.com/ouenal/scribblekitti}.} \dotfill Unknown
    \item SemanticPOSS\footnote{\url{http://www.poss.pku.edu.cn/semanticposs.html}.} \dotfill CC BY-NC-SA 3.0
    \item SemanticPOSS-API\footnote{\url{https://github.com/Theia-4869/semantic-poss-api}.} \dotfill MIT License
    \item Robo3D\footnote{\url{https://github.com/ldkong1205/Robo3D}.} \dotfill CC BY-NC-SA 4.0
\end{itemize}

\subsection{Public Implementations Used}

We acknowledge the use of the following public implementations, during the course of this work:

\begin{itemize}
    \item RangeNet++\footnote{\url{https://github.com/PRBonn/lidar-bonnetal}.} \dotfill MIT License
    \item SalsaNext\footnote{\url{https://github.com/TiagoCortinhal/SalsaNext}.} \dotfill MIT License
    \item FIDNet\footnote{\url{https://github.com/placeforyiming/IROS21-FIDNet-SemanticKITTI}.} \dotfill Unknown
    \item CENet\footnote{\url{https://github.com/huixiancheng/CENet}.} \dotfill MIT License
    \item RangeViT\footnote{\url{https://github.com/valeoai/rangevit}.} \dotfill Apache License 2.0
    \item SphereFormer\footnote{\url{https://github.com/dvlab-research/SphereFormer}.} \dotfill Apache License 2.0
    \item 2DPASS\footnote{\url{https://github.com/yanx27/2DPASS}.} \dotfill MIT License
    \item Cylinder3D\footnote{\url{https://github.com/xinge008/Cylinder3D}.} \dotfill Apache License 2.0
    \item SPVNAS\footnote{\url{https://github.com/mit-han-lab/spvnas}.} \dotfill MIT License
    \item KPConv\footnote{\url{https://github.com/HuguesTHOMAS/KPConv-PyTorch}.} \dotfill MIT License
    \item RandLA-Net\footnote{\url{https://github.com/QingyongHu/RandLA-Net}.} \dotfill CC BY-NC-SA 4.0
    \item Codes-for-PVKD\footnote{\url{https://github.com/cardwing/Codes-for-PVKD}.}\dotfill MIT License
    \item LaserMix\footnote{\url{https://github.com/ldkong1205/LaserMix}.} \dotfill Apache License 2.0
\end{itemize}

\bibliographystyle{IEEEtran}
\bibliography{ref}
\vfill

\end{document}